\documentclass[letterpaper]{article} 
\usepackage{aaai2026}  
\usepackage{times}  
\usepackage{helvet}  
\usepackage{courier}  
\usepackage[hyphens]{url}  
\usepackage{graphicx} 
\usepackage{natbib}  
\usepackage{caption} 
\usepackage{algorithm}
\usepackage{algorithmic}
\usepackage[dvipsnames]{xcolor}

\usepackage{booktabs}
\usepackage{multicol}
\usepackage{multirow}
\usepackage{colortbl}
\usepackage{xcolor}
\usepackage{caption}
\usepackage{tabularray}
\usepackage{rotating}
\usepackage{makecell}
\usepackage{tabularx}
\usepackage{float}
\usepackage{adjustbox}
\usepackage{anyfontsize}
\usepackage{mathtools}
\usepackage{amsmath}
\usepackage{indentfirst}
\usepackage{float}

\usepackage{amssymb}
\usepackage{pifont}
\newcommand{\cmark}{\ding{51}}%
\usepackage{microtype}

\usepackage{newfloat}
\usepackage{listings}
\DeclareCaptionStyle{ruled}{labelfont=normalfont,labelsep=colon,strut=off} 
\floatstyle{ruled}
\newfloat{listing}{tb}{lst}{}
\floatname{listing}{Listing}
\title{mmPred: Radar-based Human Motion Prediction in the Dark}
\author {
    Junqiao Fan\textsuperscript{\rm 1},
    Haocong Rao\textsuperscript{\rm 2},
    Jiarui Zhang\textsuperscript{\rm 1},
    Jianfei Yang\textsuperscript{\rm 1,3}\thanks{J. Yang is the project lead.},
    Lihua Xie \textsuperscript{\rm 1}\thanks{L. Xie is the corresponding author.}
}
\affiliations {
    \textsuperscript{\rm 1}School of Electrical and Electronic Engineering, Nanyang Technological University, Singapore\\
    \textsuperscript{\rm 2} College of Computing and Data Science, Nanyang Technological University, Singapore\\
    \textsuperscript{\rm 3}School of Mechanical and Aerospace Engineering, Nanyang Technological University, Singapore\\
    \text{\{fanj0019, haocong001, zhan0618\}@e.ntu.edu.sg, \{jianfei.yang, elhxie\}@ntu.edu.sg}
}

\begin{document}

\maketitle
\begin{abstract}
Existing Human Motion Prediction (HMP) methods based on RGB-D cameras are sensitive to lighting conditions and raise privacy concerns, limiting their real-world applications such as firefighting and healthcare. 
Motivated by the robustness and privacy-preserving nature of millimeter-wave (mmWave) radar, this work introduces radar as a novel sensing modality for HMP, for the first time.
%
%
Nevertheless, radar signals often suffer from specular reflections and multipath effects, resulting in noisy and temporally inconsistent measurements, such as body-part miss-detection.
To address these radar-specific artifacts, we propose mmPred, the first diffusion-based framework tailored for radar-based HMP.
mmPred introduces a dual-domain historical motion representation to guide the generation process, combining a Time-domain Pose Refinement (TPR) branch for learning fine-grained details and a Frequency-domain Dominant Motion (FDM) branch for capturing global motion trends and suppressing frame-level inconsistency. 
Furthermore, we design a Global Skeleton-relational Transformer (GST) as the diffusion backbone to model global inter-joint cooperation, enabling corrupted joints to dynamically aggregate information from others.
Extensive experiments show that mmPred achieves state-of-the-art performance, outperforming existing methods by 8.6\% on mmBody and 22\% on mm-Fi.

\end{abstract}

    
\section{Introduction}
\label{sec:introduction}

\begin{figure}[!t]
\centering
\includegraphics[width=1.\linewidth]{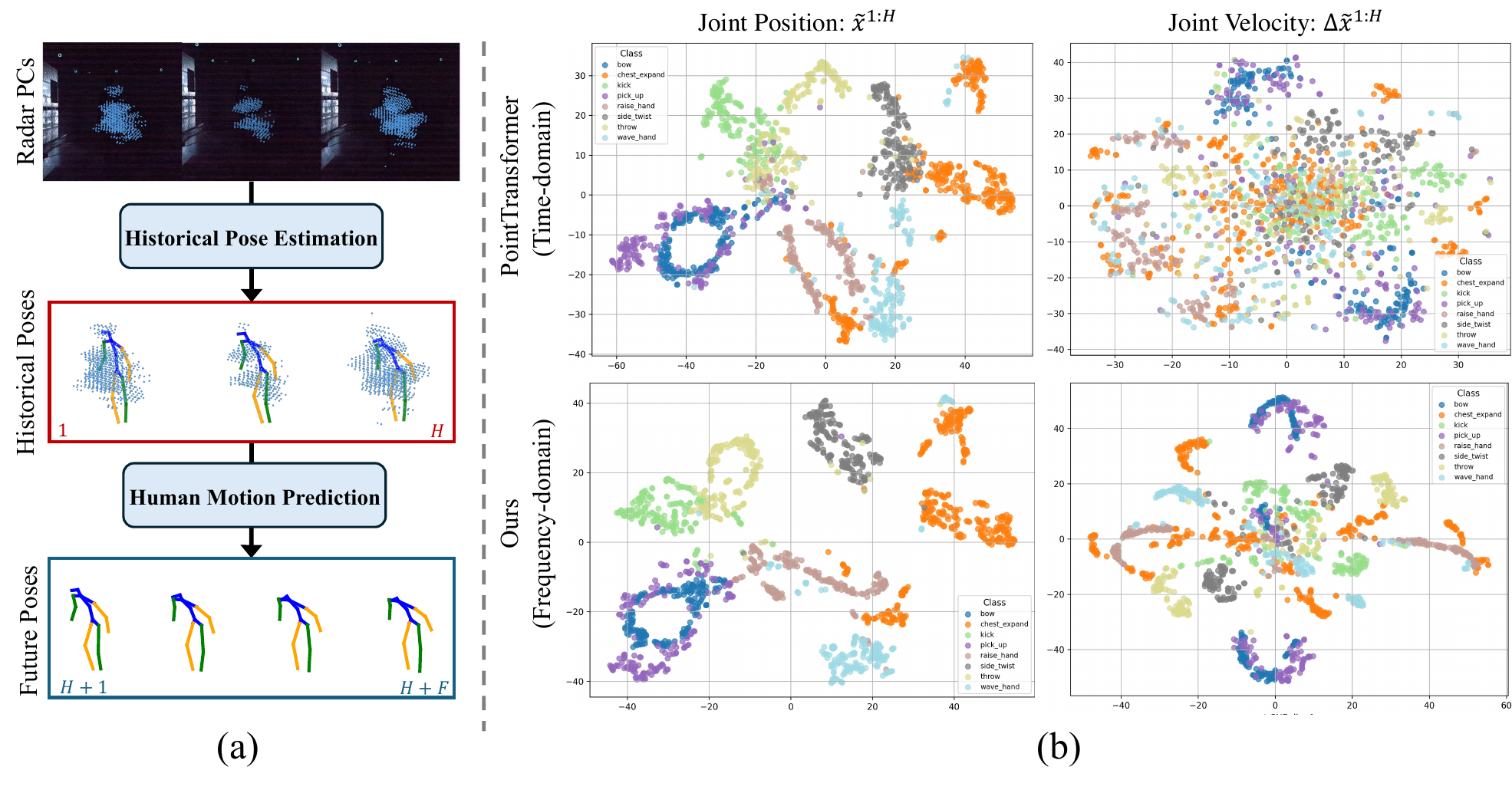}
\vspace{-2em}
\caption{(a) mmWave radar-based HMP under darkness.
(b) t-SNE visualization of joint locations and velocities from predicted historical poses. We compare our frequency-domain prediction with the state-of-the-art (SOTA) pose estimator~\cite{yang2023mm}.
Our method produces more distinguishable velocity patterns across actions than existing methods.} 

\label{Effect}
\end{figure}

%
Human Motion Prediction (HMP) aims to predict future human pose sequences from observed historical pose sequences, which plays an essential role in various applications such as human-robot interaction~\cite{gui2018teaching}, healthcare~\cite{troje2002decomposing}, and hazard prevention~\cite{yan2017development}.
Existing HMP works~\cite{li2022skeleton,chen2023humanmac} heavily rely on high-precision historical human pose sequences collected from multi-view RGBD MoCap systems, which are typically expensive and impractical for most real-world deployments.
%
Although some works~\cite{chao2017forecasting,wu2018real} attempt to use single-view RGB images, these methods struggle under adverse environmental conditions (e.g., darkness, occlusion).
Moreover, RGB-based approaches raise privacy concerns in sensitive scenarios such as elderly care.


Millimeter-wave (mmWave) radar has emerged as a promising alternative for human perception due to its low cost, portability, and scalable deployment~\cite{waldschmidt2021automotive}.
As mmWave radar operates in the 30–300 GHz frequency range, its signal can penetrate visual obscurants such as smoke and function reliably regardless of lighting conditions~\cite{zhang2023survey}.
Additionally, its limited spatial resolution naturally preserves privacy, making it suitable for HMP in indoor environments.
%
%
However, radar point clouds (PCs) are noisy due to "ghost" points caused by multipath effects~\cite{sun20213drimr}.
Moreover, intermittent miss-detections often occur due to specular reflections, where only certain body parts reflect signals back to the receiver while others deflect them away~\cite{ding2024milliflow}.
%
These issues make it challenging to capture fine-grained and realistic motion from temporally inconsistent and noisy radar PCs. 
%

To generate realistic and stable future pose sequences from radar PCs, a straightforward approach is to first estimate historical human pose sequences using an existing radar-based pose estimator, and then pass the estimated sequences into an HMP model.
However, this pipeline is difficult to succeed in practice.
Current radar-based pose estimators~\cite{chen2022mmbody, yang2023mm} often produce pose sequences with severe jitter and twisted body structures, since they process each frame independently and overlook the temporal inconsistency caused by miss-detections and multipath effect~\cite{fan2024diffusion}.
These jittery and unrealistic poses fail to preserve key motion cues, such as joint velocities (as shown in~Figure~\ref{Effect}b), which are critical for predicting future motion.
As a result, HMP models may struggle to capture meaningful temporal patterns from these radar-estimated pose sequences.
%
Moreover, existing HMP methods~\cite{martinez2017human, zhong2022spatio, li2022skeleton, chen2023humanmac} are over-sensitive to the noise of input historical pose sequences. They commonly assume clean and stable MoCap history, easily producing unrealistic future sequences under radar-specific artifacts, including ghost points and temporal inconsistency due to miss-detections.

Recently, diffusion models (DMs) have shown great potential in generating realistic human motion~\cite{tevet2022human}, through learning motion distribution and progressively removing noise.
Inspired by such capability, we propose mmPred, a diffusion-based HMP framework tailored for mmWave radar PCs.
We first introduce a dual-domain historical motion representation as complementary guidance for the diffusion model.
It captures both fine-grained pose details and dominant motion trends from noisy and temporally inconsistent radar data.
%
The Time-domain Pose Refinement (TPR) branch first estimates frame-wise human poses from radar PCs in a coarse-to-fine manner.
To mitigate the radar-specific temporal inconsistency, the Frequency-domain Dominant Motion (FDM) branch further aggregates all historical frames to extract dominant motion patterns in the frequency domain. 
By separating low-frequency motion trends from high-frequency noise, frequency-domain analysis enables the model to capture more accurate velocity patterns from noisy radar PCs, as presented in Figure~\ref{Effect}.
We then pass this guidance to a frequency diffusion model for generating future pose sequences.
To further mitigate the impact of radar-specific artifacts that may lead to unreliable guidance, we introduce the Global Skeleton-relational Transformer (GST) as the backbone of our diffusion model to enhance the realism of generated motions.
Specifically, GST employs a skeleton transformer and a frequency transformer to model joint-wise dependencies and temporal motion patterns, respectively.
In particular, the skeleton transformer enables more robust feature extraction by modeling global inter-joint cooperation, allowing unreliable joints to aggregate information from structurally or functionally related ones.
In summary, our contributions can be summarized as follows:
\begin{itemize}
    \item We propose mmPred, a novel diffusion-based HMP framework tailored for noisy and temporally inconsistent mmWave radar point clouds. To the best of our knowledge, mmPred is the first framework designed specifically for radar-based HMP.
    \item We introduce a dual-domain historical motion representation that serves as guidance for the diffusion model. It consists of a Time-domain Pose Refinement (TPR) branch to preserve fine-grained pose details, and a Frequency-domain Dominant Motion (FDM) branch to capture global motion trends while suppressing radar-specific artifacts.
    \item We design a Global Skeleton-relational Transformer (GST) as the diffusion backbone, which separately models global inter-joint cooperation and motion patterns across frequency components, further enhancing motion realism under challenging radar artifacts.
    \item Extensive experiments demonstrate that mmPred achieves state-of-the-art performance, outperforming existing methods by 8.6\% on mmBody and 22\% on mm-Fi.
\end{itemize}

\section{Related Work}
\label{sec:related_work}

\begin{figure*}[!ht]
\centering
\includegraphics[width=.9\linewidth]{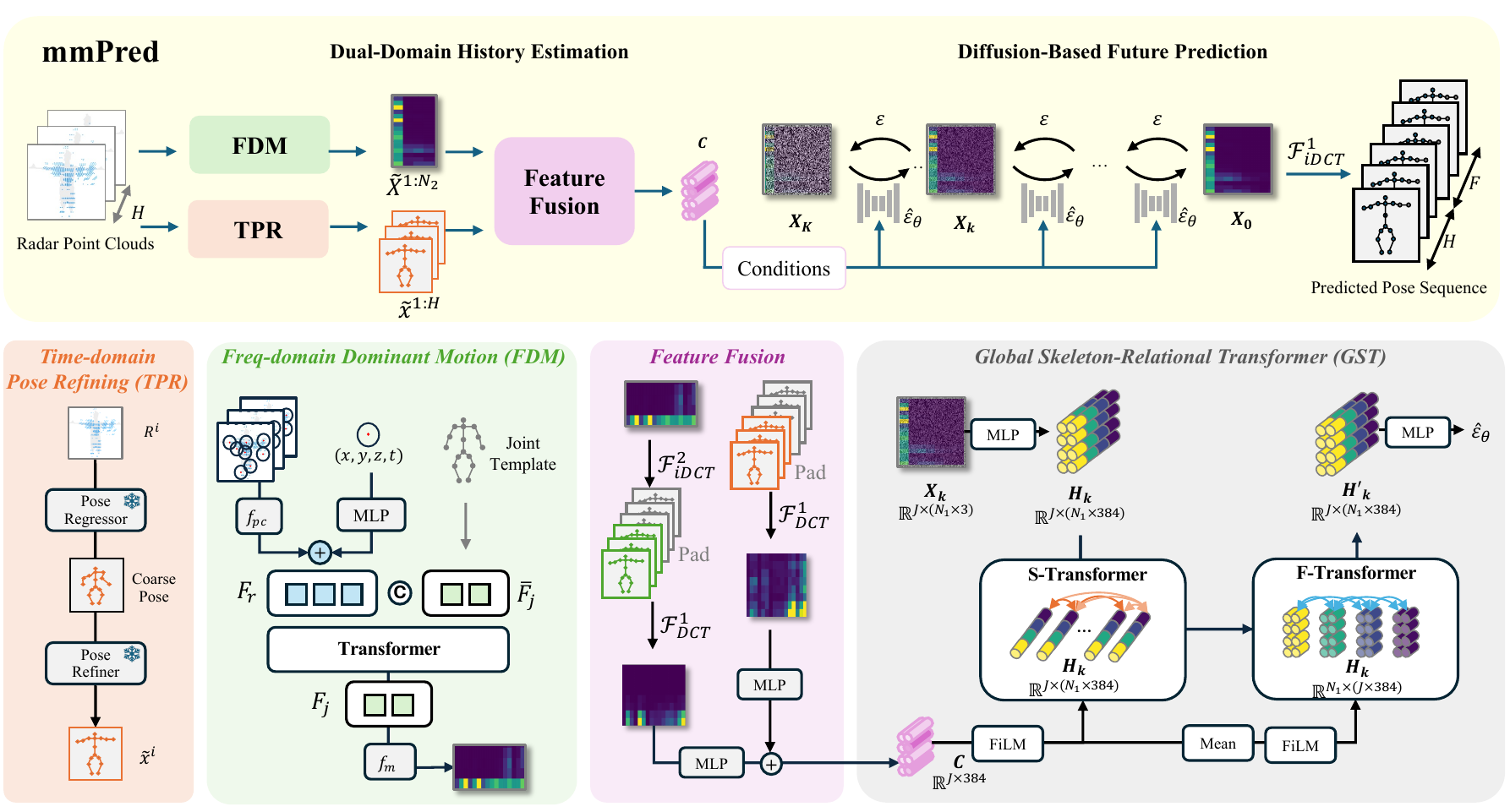}
\vspace{-0.5em}
\caption{System architecture: mmPred employs Dual-Domain History Estimation to extract historical motion representations in both the time (TPR) and frequency domains (FDM). These representations are fused via a feature fusion module to construct the condition embedding $C$, which guides the diffusion-based future motion prediction performed in the frequency domain.}
\label{system architecture}
\end{figure*}

\subsubsection{Radar-based Human Motion Sensing.}
RGB cameras have been widely used for fine-grained human sensing tasks, including 2D and 3D human pose estimation (HPE)\cite{cao2017realtime,sun2019deep,kocabas2020vibe}. However, these methods typically suffer from a severe performance drop under adverse environments~\cite{chen20232d}. Overcoming such limitations, commercial mmWave radar has emerged as a good alternative for human sensing. Early radar-based human sensing focused on coarse-level motion tracking~\cite{zhao2019mid}. More recently, mmWave radar has been explored for fine-grained HPE using radar PCs~\cite{xue2021mmmesh,an2021mars,an2022fast} or using raw radar signals~\cite{lee2023hupr,rahman2024mmvr}. Particularly, Xue et al.~\cite{xue2021mmmesh} propose an LSTM model, and Chen et al.~\cite{chen2022mmbody} and Yang et al.~\cite{yang2023mm} propose transformer-based benchmarks based on their datasets. Ding et al.~\cite{ding2024milliflow} further associate radar points with joint-level motion flow to enhance downstream human sensing tasks. Despite these advances, existing radar-based methods have not yet addressed HMP, which requires stronger modeling of spatial-temporal dynamics. These methods are typically designed in the time domain, making them vulnerable to intermittent miss-detections and failing in long-term motion understanding.

\subsubsection{Human Motion Prediction} 
Traditional HMP methods commonly employ RNNs~\cite{martinez2017human,tang2018long} and GCNs~\cite{li2022skeleton,cui2021towards,zhong2022spatio} to deterministically predict a single future, achieving reasonable accuracy but failing to capture the inherent uncertainty and multi-modality of human motion. To address this, probabilistic approaches such as DLow~\cite{yuan2020dlow} and DivSamp~\cite{dang2022diverse} introduce VAE-based frameworks, though often at the cost of generating overly diverse predictions. Recent diffusion-based methods offer a better trade-off between realism and diversity, achieving state-of-the-art performance by modeling realistic motion distributions. For example, BelFusion~\cite{barquero2023belfusion} operates in the latent space through a multi-stage training scheme, whereas HumanMAC~\cite{chen2023humanmac} adopts single-stage designs, formulating HMP as a motion inpainting task based on historical observations. Nevertheless, these HMP methods typically assume access to ideal historical poses (i.e., MoCap annotation). As a result, they are sensitive to sensor-estimated histories, especially when using radar, which provides sparse and noisy PCs that severely degrade prediction.

\section{Preliminary}
\label{sec:preliminary}


\subsubsection{Frequency-domain Motion Representation.}
\label{preliminary}
The \textit{Discrete Cosine Transform (DCT)}, a widely-used signal processing tool, can transform time-domain human pose sequences into frequency-domain coefficients.
Let $\mathbf{x}^{1:T} \in \mathbb{R}^{T \times 3J}$ be a time-domain pose sequence of ($T$) frames with ($J$) joints, and let $\mathbf{X}^{1:N} \in \mathbb{R}^{N \times 3J}$ denote the truncated DCT representation ($N \le T $), as an output of DCT. The type-II orthonormal DCT is defined as:
\begin{equation}
\mathbf{X}^{1:N} = \mathcal{F}_{DCT}(\mathbf{x}^{1:T}) = \mathbf{D} \mathbf{x}^{1:T}
\label{eq:dct_matrix},
\end{equation}
The inverse DCT reconstructs the time-domain pose sequences $\hat{\mathbf{x}}^{1:T}$ via:
\begin{equation}
\hat{\mathbf{x}}^{1:T} = \mathcal{F}_{iDCT}(\mathbf{X}^{1:N}) = \mathbf{D}^\top \mathbf{X}^{1:N}.
\label{eq:idct_matrix}
\end{equation}
In the frequency domain, dominant motion trends (e.g., mean pose and velocity) are concentrated in low-frequency components, while high-frequency components capture fine-grained details like acceleration. This spectral separation makes DCT well-suited for motion analysis under radar-specific noise.
%
\subsubsection{Frequency-domain DM for HMP.}




DM in the frequency domain directly learn the data distribution of the frequency-domain motion representation. Given the ground-truth (GT) time-domain pose sequence $\mathbf{x}^{1:(H+F)} \in \mathbb{R}^{(H+F) \times 3J}$ (with both history and future), we first apply the DCT transformation $\mathcal{F}^1_{DCT}$ using basis $\mathbf{D}_1 \in \mathbb{R}^{N_1 \times (H+F)}$ to obtain its frequency-domain representation $X^{1:N_1}$. Then, the diffusion model consists of two processes: a forward process that gradually adds noise and a reverse process that learns to invert. In the forward process, we start from the frequency-domain GT $X_0 = X^{1:N_1}$. For each diffusion step $k\in[0,..,K]$, we iteratively sample a noisier motion $X_k$ by adding Gaussian noise $\varepsilon \in \mathcal{N}(0, I)$:
\begin{equation}
    q\left(\mathbf{X}_k \mid \mathbf{X}_{k-1}\right) = \mathcal{N}\left(\mathbf{X}_k \mid \sqrt{1-\beta_k} \mathbf{X}_{k-1}, \beta_k \mathbf{I}\right),
\label{eq：forward}
\end{equation}
where $\beta_k$ is the scheduled noise scale. On the contrary, the reverse process starts from a noisy Gaussian initialization $\hat{\mathbf{X}}_K \in N(0,I)$ and progressively removes noise until $\hat{\mathbf{X}}_0$ is generated. A diffusion model, denoted as $\hat{\varepsilon}_\theta$, is trained to remove the motion noise for cleaner motion $\hat{\mathbf{X}}_{k-1}$:

\begin{equation}
    \begin{aligned}
    \hat{\mathbf{X}}_{k-1} &= (1-\beta_k)^{-1}(\hat{\mathbf{X}}_{k} - \beta_k\hat{\varepsilon}_\theta(\hat{\mathbf{X}}_k, k,C)).
    \end{aligned}
\label{eq:diff23}
\end{equation}
To ensure the predicted future motion follows historical human behavior, we leverage a condition $C$ capturing historical motion information as diffusion guidance. Finally, the output $\hat{X}_0$ from the reverse process can be transformed back to the time domain via $\mathcal{F}^1_{iDCT}$, serving as the predicted future.


\section{Methodology}
\label{sec:methodology}

\begin{figure}[!t]
\centering 
\includegraphics[width=.9\linewidth]{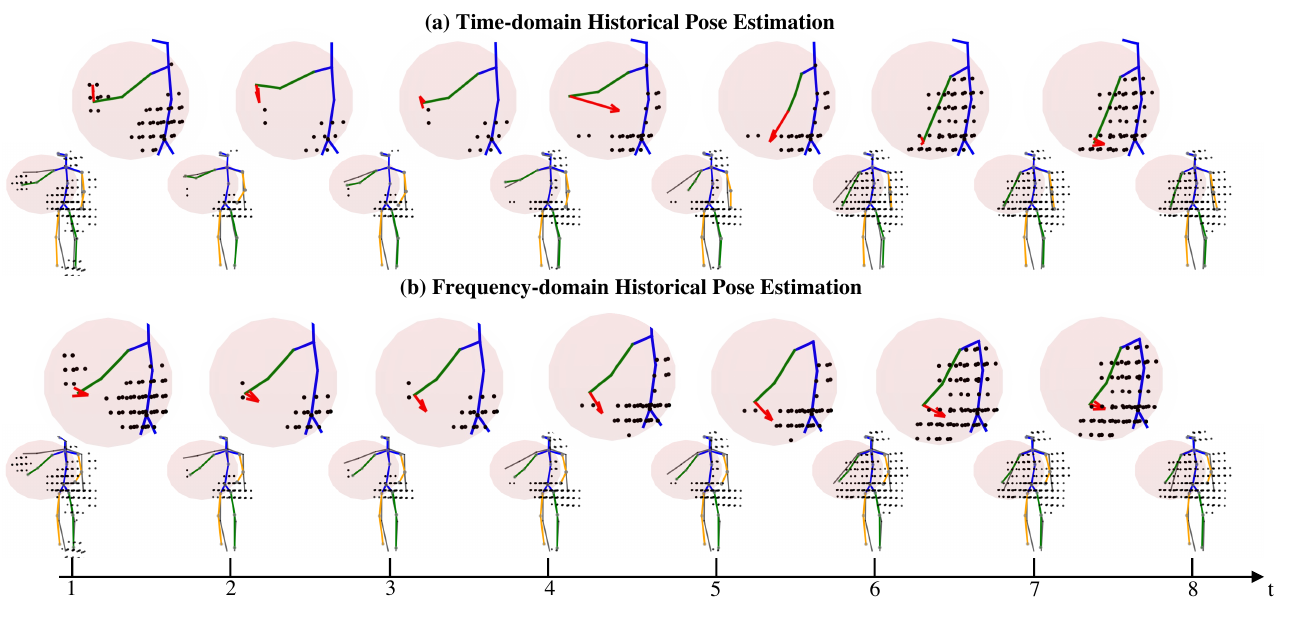}
\vspace{-0.5em}
\caption{Comparison of estimated motion history (colored) to GT (black) in dual domain. We zoom in on the right-hand area and use red arrows to mark the joint velocity. TPR demonstrates more accurate pose location, while FDM offers more consistent velocity information.}
\label{dual-ablation} 
\end{figure}

\subsection{Problem Formulation}
\label{preliminary}
Our proposed mmWave radar-based HMP task inputs historical $H$ frames of radar PCs and predicts future $F$ frames of human poses. We denote the radar PCs as $R^{1:H}=\{R^i \in \mathbb{R}^{N\times6}\}^{H}_{i=1}$, which are robust inputs under adverse environments. Each historical frame contains $N$ points, with three Cartesian coordinates and three attributes $\{v,E,A\}$ representing velocity, energy, and amplitude. The predicted future pose sequences are denoted as $\hat{x}^{H+1:H+F}=\{\hat{x}^i \in \mathbb{R}^{J\times3}\}^{H+F}_{i=H+1}$, where $J$ denotes the number of body joints. Different from the existing 3D HMP task, which takes in historical ground-truth (GT) human poses $x^{1:H}=\{x^i\}^{H}_{i=1}$, our task can only estimate historical poses from radar data, which is denoted as $\tilde{x}^{1:H}=\{\tilde{x}^i \}^{H}_{i=1}$. Given $R^{1:H}$ contains substantial noise, our task is fundamentally more challenging.

\subsection{Dual-Domain History Estimation}
\label{dual-branch}
Accurate guidance for diffusion-based motion prediction relies on robust historical motion estimation—requiring both precise joint localization and a coherent motion trend. As shown in Figure~\ref{dual-ablation}, time-domain pose estimation is easily affected by temporally inconsistent signals. To address this, we propose a dual-domain historical motion estimation approach, estimating accurate joint locations in the time domain and capturing dominant motion trends in the frequency domain.

\subsubsection{Time-domain Pose Refining (TPR).} 
For each frame $i \in [1, \dots, H]$, TPR employs a shared pretrained pose estimator~\cite{chen2022mmbody, yang2023mm}, denoted as $f_{\text{pose}}$, to predict 3D human poses from the single-frame radar point cloud $R^i$. However, occasional joint miss-detections may cause certain joints to deviate significantly from their true positions, resulting in twisted pose structures or jitter across frames. To mitigate this, we employ a pretrained diffusion-based refinement network~\cite{fan2024diffusion}, denoted as $f_{\text{refine}}$. It leverages prior knowledge such as limb-length consistency and adjacent-frame continuity to improve structural plausibility and temporal stability. Given the coarse output from $f_{\text{pose}}$, it produces refined pose estimates $\tilde{x}^i_{\text{time}}$ for each frame. Finally, we aggregate the refined poses from all $H$ frames as the TPR output:
\begin{equation}
    \begin{aligned}
    \tilde{x}_{\text{time}}^{i} &= f_{\text{refine}}(f_{\text{pose}}(R^i))\\
    \tilde{x}_{\text{time}}^{1:H} &= \{\tilde{x}_{\text{time}}^{1},\tilde{x}_{\text{time}}^{2},...,\tilde{x}_{\text{time}}^{H}\}^H_{i=1} 
    \end{aligned}
\end{equation}

\subsubsection{Frequency-domain Dominant Motion (FDM).} 
FDM directly outputs a compact frequency-domain motion representation, treating the entire historical motion sequence holistically. It takes all $R^{1:H}$ as input and predicts the frequency-domain motion representation, $\tilde{X}^{1:N_2}_{\text{freq}}\in \mathbb{R}^{N_2\times J \times 3}$, where $N_2$ denotes the number of selected coefficients (typically $N_2 = 3$ or $4$). These few coefficients capture the most dominant motion dynamics, allowing the model to focus on a consistent motion trend without sudden changes. 

As shown in Figure~\ref{system architecture}, we first capture the global evolution of PCs into a feature representation. The $R^{1:H}$ is encoded by anchor-based PC encoder$f_{\text{pc}}$~\cite{zhao2021point}. Specifically, for each frame, we select $N'$ representative anchor points using Farthest Point Sampling (FPS). The selected anchors encode local PC features from the neighborhood, aggregated into anchor features $F_{r}\in\mathbb{R}^{(H \times N') \times 1024}$. To encode the evolution of PCs over time, we inject temporal information using an MLP that maps each anchor’s $(x, y, z, t)$ (where $t \in [1, H]$) to a positional embedding, which is then added to the corresponding anchor features. 
Then, we devise a transformer-based architecture to perform frequency-domain motion learning. A transformer $\Phi$ is applied to dynamically project time-domain anchor features into frequency-domain joint motion features. We first zero-initialize a set of learnable joint template tokens $\bar{F}_j \in \mathbb{R}^{J \times 1024}$, one for each body joint. These are concatenated with the anchor features and jointly processed by $\Phi$:
\begin{equation}
    \begin{aligned}
        F_r &= f_{\text{pc}}(R^{1:H}),\\
        F'_r, F_j &= \Phi(F_r, \bar{F}_j).
    \end{aligned}   
\end{equation}
Within the transformer, deep correlation is captured, both within $F_r$ and between $F_j$ and $F_r$. The $F_r$ captures the temporal PC evolution through anchor-wise self-attention. Each joint feature in $F_j$ also dynamically aggregates information from correlated anchor feature cues. The captured joint features $F_j$ are eventually decoded by an MLP-based motion decoder $f_{m}$ to produce the frequency-domain representation:
\begin{equation}
    \tilde{X}_{\text{freq}}^{1:N_2} = f_{m}(F_j).
\end{equation}

\subsubsection{Cross-Domain History Fusion.} 

To guide the diffusion model, we fuse the dual-domain  $\tilde{x}_{time}^{1:H}$ and $\tilde{X}_{freq}^{1:N_2}$ into a conditional embedding $C$. Since the diffusion model operates in the frequency domain, we transform both inputs into frequency domain representations, $\tilde{X}_{\text{time}}^{1:N_1}$ and $\tilde{X}_{\text{freq}}^{1:N_1}$, with $N_1$ DCT coefficients corresponding to $H+F$ frames. Specifically, both representations are first converted to the time domain to perform repeated padding up to $H + F$ frames, and then transformed back into the frequency domain:

\begin{equation}
\tilde{X}_{\text{time}}^{1:N_1} = \mathcal{F}_{DCT}^1(\text{pad}(\tilde{x}_{\text{time}}^{1:H})),
\end{equation}

\begin{equation}
\tilde{X}_{\text{freq}}^{1:N_1} = \mathcal{F}_{DCT}^1\left(\text{pad}\left(\mathcal{F}_{iDCT}^2(\tilde{X}_{\text{freq}}^{1:N_2})\right)\right).
\end{equation}
Both representations are then reshaped into $\mathbb{R}^{J\times(N_1 \times 3)}$, isolating each joint to preserve its distinct motion dynamics and prevent interference from unreliable keypoints caused by radar miss-detections. Then, the reshaped representations are independently projected via two MLPs $f_1$ and $f_2$, and fused by element-wise addition to obtain the final joint-wise conditional embedding $C\in\mathbb{R}^{J\times 384}$:
\begin{equation}
    C = f_1(\tilde{X}_{\text{time}}^{1:N_1}) + f_2(\tilde{X}_{\text{freq}}^{1:N_1}). 
\end{equation}

\begin{figure*}[!t]
\centering
\includegraphics[width=.75 \linewidth]{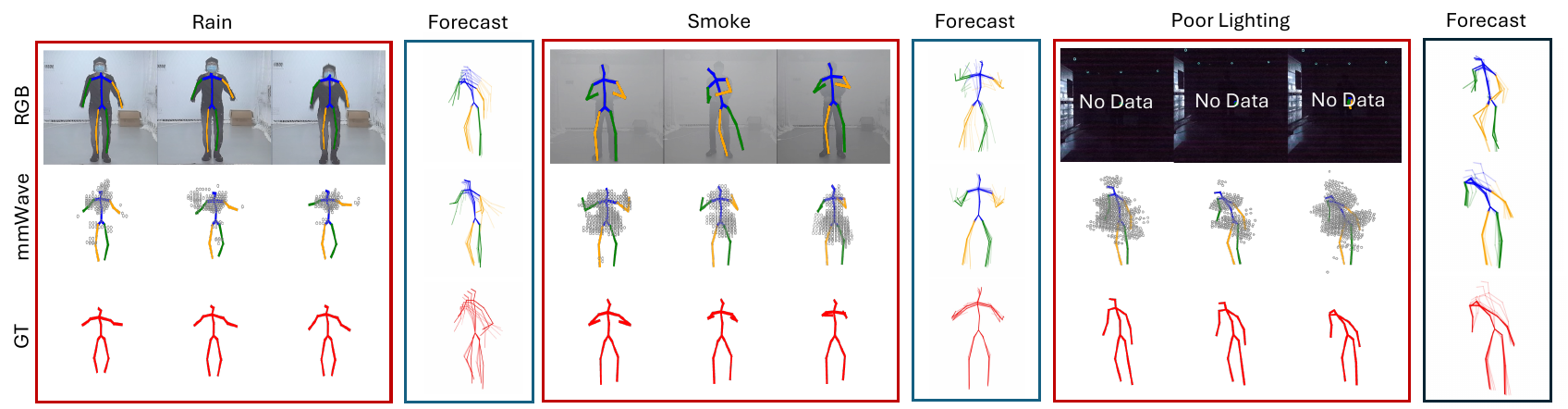}
\vspace{-0.5em}
\caption{HMP under adverse environments using RGB, mmWave (ours), and MoCap GT. The red boxes enclose estimated historical poses, and the blue boxes enclose the predicted future poses (predicted frames are stacked).}
\label{exp1_effect} 
\end{figure*}

\subsection{Global Skeleton-Relational Transformer (GST)}
\label{spatial-temporal}

Existing transformer-based diffusion models typically overlook global cooperation among joints by encoding all joints into a unified representation, making them vulnerable to joint miss-detections.
To address this, GST first isolates joint-specific features to preserve individual motion dynamics, then applies joint-wise self-attention to enable global information exchange. As shown in Figure~\ref{system architecture}, the noisy motion of intermediate diffusion step $k$, $X_k \in \mathbb{R}^{N_1 \times J \times 3}$, is first projected to latent features $H_k \in \mathbb{R}^{N_1 \times J \times 384}$ by an MLP. This yields $N_1 \times J$ tokens, one for each joint at each frequency coefficient.
For spatial modeling, we reshape $H_k$ into $\mathbb{R}^{J \times (N_1 \times 384)}$, aggregating all frequency components into joint-specific tokens. These $J$ tokens are passed through a Skeleton Transformer (S-Transformer), where self-attention enables joint-wise interaction. Potentially, mis-detected joints recover meaningful representations by leveraging information from other observed joints. 
Next, for temporal modeling, $H_k$ is reshaped into $\mathbb{R}^{N_1 \times (J \times C_a)}$, forming $N_1$ tokens, each aggregating information across all joints at a given frequency. These tokens are processed by a Frequency Transformer (F-Transformer), which applies self-attention along the temporal dimension to ensure the generated motion sequence is temporally smooth and realistic.

To inject the conditional information, we adopt the Feature-wise Linear Modulation (FiLM) strategy. For the S-Transformer, we maintain joint-wise conditioning by directly inputting $C \in \mathbb{R}^{J \times 384}$, allowing each joint to receive individualized control. For the F-Transformer, which processes joint-aggregated tokens, we apply mean-pooling over the joint dimension to obtain a global condition $C' \in \mathbb{R}^{1 \times 384}$ before applying FiLM.

\subsection{Overall Learning Objective}
\label{overall training}

The Dual-Domain Historical Motion Estimation and the Diffusion-based Future Prediction are trained in two separate stages. Firstly, the FDM module is trained using an $L_2$ loss with the GT frequency-domain motion as supervision:
\begin{equation}
    \mathcal{L}_{1}=||\tilde{X}^{1:N_2}-\mathcal{F}_{DCT}^{2}(x^{1:H})||^2.
\end{equation}
Subsequently, with the pretrained TPR and FDM, we train the GST-based diffusion model using the $\varepsilon$-prediction objective from Denoising Diffusion Probabilistic Models (DDPM)~\cite{ho2020denoising}, which minimizes the error between the predicted and true noise added at each diffusion timestep.
\begin{equation}
    \mathcal{L}_{2}=||\hat{\varepsilon}_{\theta}(X_k, k, C) - \varepsilon_k||^2.
\end{equation}

\section{Experiments}
\label{sec:experiments}

\begin{table*}[!t]
\footnotesize
\centering
\small
\setlength{\tabcolsep}{1.5pt}  


    \begin{tabular}{c|l|cc|cc|cc|cc|cc|cc|cc|cc}
    \toprule
    \multicolumn{2}{c|}{\multirow{2}[2]{*}{Methods}} & \multicolumn{2}{c|}{Lab1} & \multicolumn{2}{c|}{Lab2} & \multicolumn{2}{c|}{Furnished} & \multicolumn{2}{c|}{Rain} & \multicolumn{2}{c|}{Smoke} & \multicolumn{2}{c|}{Dark} & \multicolumn{2}{c|}{Occlusion} & \multicolumn{2}{c}{Avg.} \\
    \multicolumn{2}{c|}{} & ADE\textdownarrow & FDE\textdownarrow & ADE\textdownarrow & FDE\textdownarrow & ADE\textdownarrow & FDE\textdownarrow & ADE\textdownarrow & FDE\textdownarrow & ADE\textdownarrow & FDE\textdownarrow & ADE\textdownarrow & FDE\textdownarrow & ADE\textdownarrow & FDE\textdownarrow & ADE\textdownarrow & FDE\textdownarrow \\
    \midrule
    \midrule
    GT    & HumanMAC & 0.235 & 0.329 & 0.242 & 0.344 & 0.278 & 0.370 & 0.297 & 0.380 & 0.338 & 0.462 & 0.287 & 0.394 & 0.265 & 0.358 & 0.291 & 0.392 \\
    RGB   & HumanMAC & 0.390 & 0.452 & 0.427 & 0.516 & 0.420 & 0.482 & 0.479 & 0.501 & 0.560 & 0.589 & 0.693 & 0.612 & 0.739 & 0.627 & 0.547 & 0.550 \\
    \midrule
    \multirow{3}[2]{*}{mmWave} & PSGSN & 0.503 & 0.597 & 0.526 & 0.640 & 0.524 & 0.626 & 0.536 & 0.580 & 0.598 & 0.684 & 0.513 & 0.600 & 0.485 & 0.566 & 0.537 & 0.617 \\
          & HumanMAC & 0.411 & 0.458 & 0.441 & 0.505 & 0.450 & 0.478 & 0.455 & 0.465 & 0.496 & 0.540 & 0.406 & 0.447 & 0.391 & 0.430 & 0.460 & 0.487 \\
          & mmPred & \textbf{0.369} & \textbf{0.412} & \textbf{0.387} & \textbf{0.441} & \textbf{0.418} & \textbf{0.439} & \textbf{0.436} & \textbf{0.444} & \textbf{0.472} & \textbf{0.522} & \textbf{0.392} & \textbf{0.436} & \textbf{0.378} & \textbf{0.416} & \textbf{0.420} & \textbf{0.456} \\
    \bottomrule
    \end{tabular}%

\caption{Quantitative results on the mmBody under different adversarial environments, evaluated using ADE and FDE metrics. We compare different input modalities for historical motion, including RGB and mmWave radar. GT-based methods are provided to indicate the upper-bound performance. Bold indicates the best performance among sensor-derived methods.}
    \label{tab:maintable}
    
\end{table*}%

\begin{figure*}[!t]
\centering
\includegraphics[width=.8\linewidth]{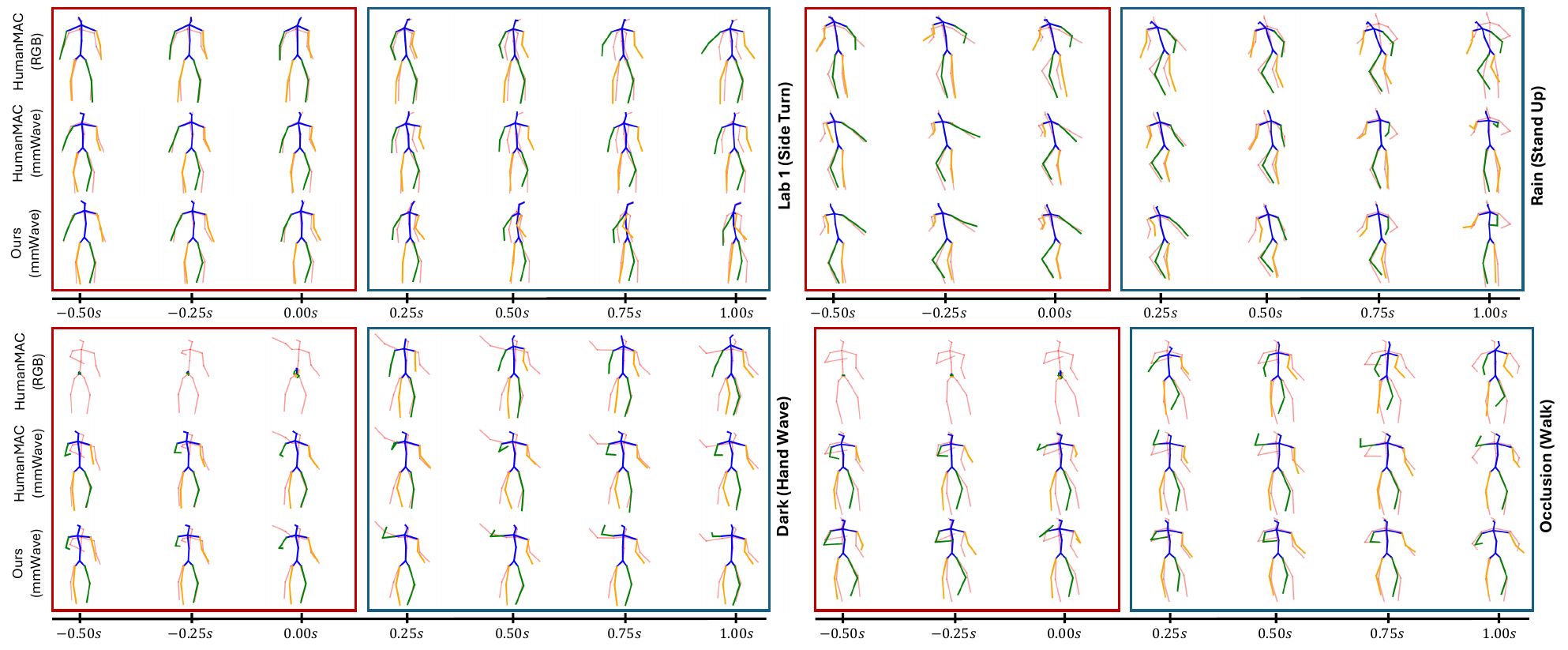}
\vspace{-0.5em}
\caption{Qualitative visualization of our method compared to baseline on the mmBody dataset, where red poses denote GT and colored poses denote our prediction. Red boxes enclose the 0.5s historical frames, and blue boxes enclose the 1s predicted future. The higher overlap between predicted and gt poses indicates higher accuracy.}
\label{exp1} 
\end{figure*}

\subsection{Experimental Setup}
\label{Experiment Setup}

\subsubsection{Datasets.}
Our experiments are conducted on mmBody~\cite{yang2023mm} and mm-Fi~\cite{yang2023mm} datasets. Specifically, mmBody collects training data under normal scenes (Lab1 and Lab2) and testing data under adverse environments. It contains MoCap-annotated poses and multi-modal sensor data from RGB-D and Phenix mmWave radar. This setting enables robustness comparison of different modalities. We select $H=8$ frames (0.5s) for history and $F=16$ frames (1s) for future prediction. Mm-Fi is a larger-scale human motion dataset using a lower-bandwidth mmWave radar with sparser PCs. The human poses are annotated from RGB images using the pretrained HRNet-w48~\cite{sun2019deep}. We adopt a random train-test split following the daily-activity protocol. Due to distinct frame rates, we select $H=5$ frames (0.5s) for history and $F=10$ frames (1s) for future prediction.

\subsubsection{Competing Methods.} 
To adapt to raw sensor data input, we adopt a two-stage protocol for all evaluated methods: historical pose estimation followed by future motion prediction. During training and testing, HMP methods accepted historical poses generated by sensor-specific pretrained pose estimators. The output of HMP models is compared with the GT annotations. This setting reflects realistic deployment scenarios, where only sensor data is available.
We compare our method with two representative SOTA open-source HMP models: a graph-based model, PSGSN~\cite{li2022skeleton}, and a diffusion-based model, HumanMAC~\cite{chen2023humanmac}. For the pose estimators, we follow the benchmark practices on different datasets. On mmBody, we use VIBE\cite{kocabas2020vibe} for RGB video and P4Transformer\cite{fan2021point} for radar PCs. On mm-Fi, we adopt PointTransformer~\cite{zhao2021point} for radar PCs. The estimator-generated pose sequences are then used as input to train and evaluate all HMP models.


\subsubsection{Evaluation Metric.}
Following previous HMP works~\cite{chen2023humanmac}, we adopt two metrics designed for future motion accuracy: (1) Average Displacement Error (ADE), the mean L2 distance between predictions and GT over all future frames; and (2) Final Displacement Error (FDE), the L2 distance at the final frame. Following the best-of-K evaluation for diffusion-based methods, we generate 10 hypotheses and report the best performance. Full assessments of model diversity and multimodal accuracy are presented in Tab.~\ref{tab:mmModality}.

\begin{table*}[t]
\footnotesize
\centering
\small
\setlength{\tabcolsep}{1.5pt}  

    \begin{tabular}{c|l|cc|cc|cc|cc|cc|cc|cc|cc}
    \toprule
    \multicolumn{2}{c|}{\multirow{2}[2]{*}{Method}} & \multicolumn{2}{c|}{Chest Expand} & \multicolumn{2}{c|}{Side Twist} & \multicolumn{2}{c|}{Raise Hand} & \multicolumn{2}{c|}{Pickup} & \multicolumn{2}{c|}{Throwing} & \multicolumn{2}{c|}{Kicking} & \multicolumn{2}{c|}{Bowing} & \multicolumn{2}{c}{Avg.} \\
    \multicolumn{2}{c|}{} & ADE\textdownarrow   & FDE\textdownarrow   & ADE\textdownarrow   & FDE\textdownarrow   & ADE\textdownarrow   & FDE\textdownarrow   & ADE\textdownarrow   & FDE\textdownarrow   & ADE\textdownarrow   & FDE\textdownarrow   & ADE\textdownarrow   & FDE\textdownarrow   & ADE\textdownarrow   & FDE\textdownarrow   & ADE\textdownarrow   & FDE\textdownarrow \\
    \midrule
    \midrule
    GT    & HumanMAC & 0.300 & 0.256 & 0.277 & 0.276 & 0.244 & 0.226 & 0.414 & 0.486 & 0.374 & 0.393 & 0.337 & 0.366 & 0.251 & 0.256 & 0.293 & 0.293 \\
    \midrule
    \multirow{3}[2]{*}{mmWave} & PSGSN & 0.402 & 0.422 & 0.430 & 0.445 & 0.397 & 0.446 & 0.595 & 0.686 & 0.449 & 0.517 & 0.426 & 0.449 & 0.336 & 0.395 & 0.430 & 0.470 \\
          & HumanMAC & 0.353 & 0.306 & 0.407 & 0.392 & 0.363 & 0.333 & 0.547 & 0.578 & 0.439 & 0.471 & 0.425 & 0.421 & 0.304 & 0.314 & 0.408 & 0.396 \\
          & mmPred & \textbf{0.272} & \textbf{0.230} & \textbf{0.315} & \textbf{0.308} & \textbf{0.237} & \textbf{0.218} & \textbf{0.452} & \textbf{0.481} & \textbf{0.371} & \textbf{0.370} & \textbf{0.374} & \textbf{0.368} & \textbf{0.246} & \textbf{0.234} & \textbf{0.319} & \textbf{0.305} \\
    \bottomrule
    \end{tabular}%
\caption{Quantitative results on mm-Fi for different actions, evaluated by ADE and FDE. Noted that mm-Fi adopts GT from RGB images using the pretrained HRNet-w48~\cite{sun2019deep}. Bold indicates the best performance among sensor-driven methods.}
    \label{tab:mmfi}
\end{table*}%

\begin{figure*}[t]
\centering
\includegraphics[width=.95\linewidth]{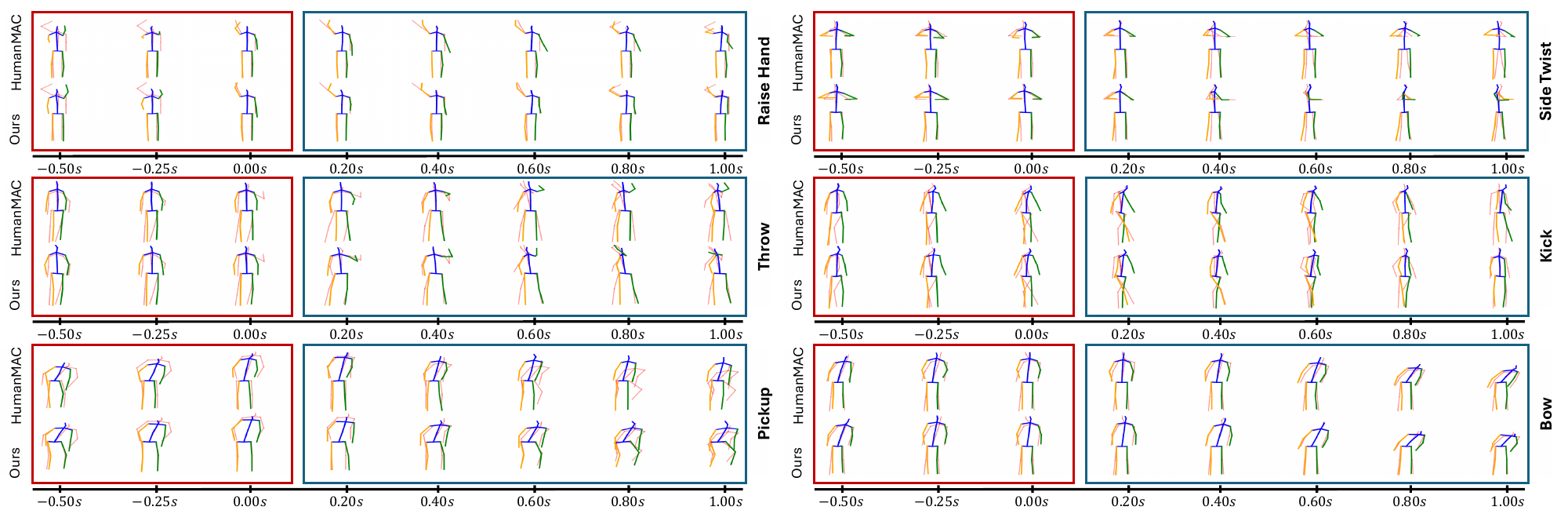}
\vspace{-0.5em}
\caption{Qualitative visualization of our method compared to baseline on the mm-Fi dataset.}
\label{exp2}
\end{figure*}

\begin{table}[!b]

\setlength{\tabcolsep}{3 pt}  
\centering

\small  

    \begin{tabular}{c|ccc|cc|cc}
    \toprule
    \multirow{2}[2]{*}{Methods} & \multicolumn{3}{c|}{Modules} & \multicolumn{2}{c|}{mmBody} & \multicolumn{2}{c}{mm-Fi} \\
          & TPR   & FDM   & GST   & ADE\textdownarrow   & FDE\textdownarrow   & ADE\textdownarrow   & FDE\textdownarrow \\
    \midrule
    M1    &       &       &       & 0.460 & 0.487 & 0.408 & 0.396 \\
    M2    & \cmark     &       &       & 0.455 & 0.485 & 0.379 & 0.359 \\
    M3    &       & \cmark     &       & 0.456 & 0.486 & 0.337 & 0.326 \\
    M4    &       &       & \cmark     & 0.460 & 0.485 & 0.373 & 0.354 \\
    \midrule
    M5    & \cmark     & \cmark   &     & 0.448 & 0.476 & 0.355 & 0.327 \\ 
    M6    &       & \cmark     & \cmark   & 0.423 & 0.458 & 0.325 & 0.310 \\  
    M7    & \cmark     &       & \cmark   & 0.439 & 0.474 & 0.334 & 0.324 \\
    M8    & \cmark     & \cmark     & \cmark     & \textbf{0.420} & \textbf{0.456} & \textbf{0.319} & \textbf{0.305} \\
    \bottomrule
    \end{tabular}%

\caption{Ablation studies of proposed modules on mmBody and mm-Fi. Specifically, M2, M3, and M4 show the efficiency of TPR, FDM, and GST, respectively. M5, M6, and M7 explore the necessity of GST, TPR, and FDM, respectively.}
\label{tab:ablation}
\end{table}

\begin{table}[!b]
\centering
\small
\setlength{\tabcolsep}{2 pt}

    \begin{tabular}{l|ccp{5pt}c|cc}
\cmidrule{1-3}\cmidrule{5-7}          & \multicolumn{2}{c}{Realism} &       &    Feat.   & \multicolumn{2}{c}{Accuracy} \\
    GST   & Error\textdownarrow & Jitter\textdownarrow &       & Isolation & ADE\textdownarrow   & FDE\textdownarrow \\
\cmidrule{1-3}\cmidrule{5-7}    w/o S-Tran. & 10.67 & 7.01 &       & no    & 0.438 & 0.470 \\
    w/ S-Trans. & \textbf{9.92} & \textbf{6.09} &       & yes   & \textbf{0.420 }& \textbf{0.456} \\
\cmidrule{1-3}\cmidrule{5-7}    \end{tabular}%
\caption{Ablation study of the GST module. \textbf{Left:} Effect of the S-Transformer on realism metrics, including limb-length error and limb-length jitter. \textbf{Right:} Effect of GST's joint-wise isolation design on prediction accuracy.}
\label{tab:realism-ablat}
\end{table}

\subsection{Overall Result}
\label{Overall Result}
\subsubsection{Comparison with RGB-based HMP.}
As presented in Table~\ref{tab:maintable}. Due to sensor noise and inaccuracies in historical pose estimation, methods based on raw sensor input generally underperform those using GT history (upper-bound performance). However, mmWave-based HMP methods demonstrate superior robustness across all adverse environments. Notably, mmPred reduces ADE and FDE by over 9\% and 11.4\% under rain and smoke, and improves ADE by more than 40\% under darkness and occlusion. This is largely because RGB-based methods fail in low-light or occluded scenarios, often producing random outputs. As visualized in Figure~\ref{exp1_effect}, RGB inputs degrade severely under challenging conditions, leading to erroneous historical poses and inaccurate predictions, whereas mmWave radar maintains reliable sensing, enabling accurate motion forecasting.

\subsubsection{Performance on mmBody.} 
Compared to existing methods leveraging radar PCs, mmPred achieves significantly better performance. As shown in Figure~\ref{exp1}, radar estimated pose sequences often exhibit jitter and erroneous velocities, while prior approaches are highly sensitive to these radar-specific noises. Traditional GCN-based encoder-decoder models like PSGSN struggle to handle such noise. While diffusion-based methods like HumanMAC offer improvements, they remain vulnerable to corrupted historical conditions. In contrast, mmPred outperforms HumanMAC by an average of 8.6\% in ADE and 6.4\% in FDE across various adverse conditions. This improvement is largely attributed to the proposed FDM and the GST-based diffusion architecture. FDM captures more stable motion trends and velocity cues, while GST mitigates the impact of unreliable joints through joint-wise cooperation. As illustrated in Figure~\ref{exp1}, baseline methods are easily misled by jitter in historical poses, particularly for hands, resulting in incorrect motion direction and velocity estimation.

\subsubsection{Performance on mm-Fi.} 
As shown in Table~\ref{tab:mmfi}, we further evaluate the generalizability of mmPred on the mm-Fi dataset, typically with sparse PCs and more severe pose jitter in history. Under these challenges, mmPred achieves a more significant improvement over existing methods. Specifically, it reduces ADE and FDE by 22\% and 23\%, respectively, compared to the SOTA HumanMAC. These results highlight that existing diffusion-based HMP methods are sensitive to historical noise, whereas mmPred remains robust through frequency-domain motion modeling. Qualitatively, as presented in Figure~\ref{exp2}, HumanMAC produces inaccurate predictions, while mmPred yields trajectories that better align with the GT. In motions like side twist and pickup, pose jitter disrupts slight motion trends, making them harder to predict. FDM mitigates this by extracting global cues from radar flow, offering more reliable temporal guidance. For larger motions such as throwing and kicking, GST enhances joint-wise coordination, yielding more anatomically consistent future poses.

\subsection{Ablation Study}
\label{Analytics}

In Table~\ref{tab:ablation}, we present ablation studies on the three proposed modules in mmPred: TPR, FDM, and GST. All modules contribute to reducing future prediction errors, with FDM showing particularly strong gains on the mm-Fi dataset, highlighting its effectiveness under extreme sparsity. Additionally, experiments M5–M7 demonstrate performance drops when removing any single module, confirming their necessity for achieving optimal results. Further ablations on the S-Transformer and feature isolation design are reported in Table~\ref{tab:realism-ablat}. Following~\cite{curreli2025nonisotropic}, we adopt two metrics to quantify motion realism: (1) limb-length error, the normalized $L_1$ distance between predicted and ground-truth limb lengths; and (2) limb-length jitter, the normalized $L_1$ distance between predicted limb lengths across consecutive frames. Results show that the S-Transformer significantly reduces both metrics, while the feature isolation design further enhances accuracy by preventing occasional keypoint miss-detections from corrupting the unified feature representation.

\section{Conclusion}\label{sec:conclusion}
This paper presents mmPred, the first diffusion-based HMP framework tailored for the mmWave radar modality. To address the challenges posed by noisy and unstable radar signals, we propose a dual-domain historical motion representation as the diffusion condition. Furthermore, the Global Skeleton Transformer backbone enhances motion realism by modeling joint-wise correlations. Extensive experiments demonstrate that mmPred achieves superior robustness and accuracy under adverse conditions compared to existing methods. In the future, cross-domain generalization and the incorporation of raw radar signals warrant further exploration.

\section*{Acknowledgements}
This work is supported by Ministry of Education (MOE), Singapore, under AcRF TIER 1 Grant RG64/23 and National Research Foundation of Singapore Medium-sized Centre for Advanced Robotics Technology Innovation. This work is jointly supported by MOE Singapore Tier 1 Grant RG83/25, RS36/24 and a Start-up Grant from Nanyang Technological University.

\clearpage
\setcounter{section}{0}

\section{Content of the Appendix Materials}
The appendix is organized as follows: (1) an overview of mmWave radar-based human sensing principles; (2) details on the construction of our HMP datasets; (3) supplementary implementation details of the proposed mmPred framework; and (4) additional experimental results.





\section*{Appendix}
\section{mmWave Radar-based Human Sensing}
\begin{figure}[!b]
\centering
\includegraphics[width=3.0in]{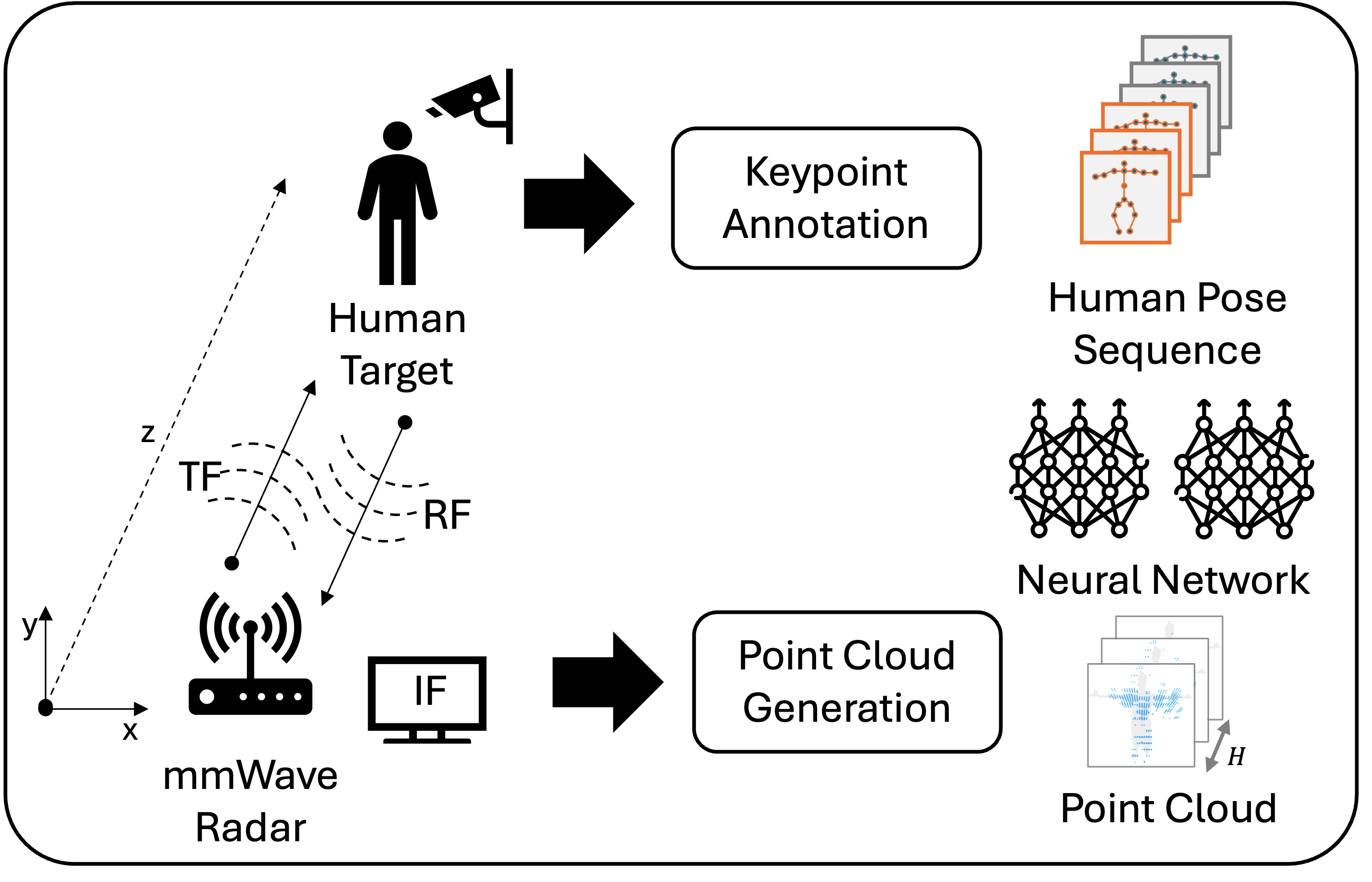}
\caption{Pipeline for mmWave radar-based human sensing and Human Motion Prediction (HMP).}
\label{sensing}
\end{figure}

In Figure~\ref{sensing}, we illustrate the pipeline of mmWave radar for human sensing, which detects human actions within a range of 3-5 meters from the radar. This process generates mmWave point clouds (PCs). Simultaneously, a keypoint annotation system such as VICON, Mocap, or Cameras is deployed to record ground-truth human poses for reference. The mmWave radar-based human motion prediction (HMP) refers to training a neural network to first estimate historical human pose sequences and predict future pose sequences, using mmWave radar PCs as input.

For general mmWave human sensing, FMCW (Frequency Modulated Continuous Wave) chirp signals are transmitted and their reflections are received through antenna arrays. These chirp signals are defined by parameters such as start frequency $f_c$, bandwidth $B$, and duration $T_c$. To generate radar PCs~\cite{xue2021mmmesh}, range-FFT separates different frequency components $f$ from the IF signals, enabling the extraction of object distances using the formula $R=\frac{cfT_c}{2B}$, where $c$ is the speed of light. Doppler-FFT measures phase changes $\omega$ of the IF signals, facilitating the calculation of object velocities using $v=\frac{\lambda\omega}{4\pi T_c}$, where $\lambda$ is the wavelength of the chirp. Elevation angles $\varphi$ and azimuth angles $\theta$ of the detected objects are determined based on $\varphi=sin^{-1}(\frac{\omega_z}{\pi})$ and $\theta=sin^{-1}(\frac{\omega_x}{cos(\varphi)\pi})$, where $\omega_x$ is the phase change between azimuth antennas and corresponding elevation antennas, and $\omega_z$ is the phase change of consecutive azimuth antennas. Finally, the Cartesian coordinates (x, y, z) of the detected point clouds are calculated as follows: $x=R cos(\varphi) sin(\theta)$, $z=R sin(\varphi)$, and $y=(R^2-x^2-z^2)$.

\section{Construction of mmWave Radar-based HMP Datasets}

\subsection{Datasets}

\begin{table*}[!t]
\footnotesize
\centering
\caption{Overview of the datasets used for mmWave human pose estimation. }
\setlength{\belowcaptionskip}{0.2cm}
\resizebox{1.\textwidth}{!}{
  \fontsize{12}{14}\selectfont
\begin{tabular}{p{8em} p{11.5em} p{1em} p{16.5em} p{1em} p{13.5em}}
    \toprule
    \textbf{Dataset:} & \textbf{mmMesh (Not Public)} & & \textbf{mmBody (Public)} & & \textbf{mm-Fi (Public)} \\
    \midrule[1.5pt] \midrule
    \textbf{Radar Type:} & AWR1843BOOST mmWave radar from Texas Instruments. & & Phoenix mmWave radar from Arbe Robotics (15fps). & & IWR6843 60-64GHz mmWave radar from Texas Instruments (10fps). \\
    \midrule
    \textbf{Annotations:} & Mesh annotated by VICON motion capture system and generated by SMPL. & & 55 keypoints are annotated by the OptiTrack Mocap system; Mesh is generated by Mosh++ and SMPL-X. & & 2D keypoints are obtained by HRNet-w48 from two-view infra-red cameras; 3D keypoints are calculated by triangulation. \\
    \midrule
    \textbf{Point format:} & Cartesian: (x, y, z); PC attributes: (range, velocity, energy). & & Cartesian: (x, y, z); PC attributes: (velocity, amplitude, energy). & & Cartesian: (x, y, z); PC attributes: (velocity, intensity).\\
    \midrule
    \textbf{\# of subjects:} & Not mentioned. & & 20 (10 males, 10 females). & & 40 (29 males and 11 females) \\
    \midrule
    \textbf{\# of actions:} & Not mentioned. & & 100 motions (16 static poses, 9 torso motions, 20 leg motions, 25 arm motions, 3 neck motions, 14 sports motions, 7 daily indoor motions, and 6 kitchen motions). & & 27 actions (14 daily activities and 13 rehabilitation exercises) for a duration of 30 seconds. \\
    \midrule
    \textbf{\# of samples:} & Not mentioned. & & 7495 frames for training and 5534 frames for testing. & & 12096 frames for training and 3024 frames for testing. \\
    \midrule
    \textbf{Scenes:} & Normal and occlusion. & & Lab1, Lab2, Furnished, Poor\_lighting, Rain, Smoke, and Occlusion.  & & Normal, Cross-subject, and Cross-environment. \\
    \bottomrule
    \end{tabular}
    }
    
    \label{tab:dataset}
\end{table*}%

As shown in Table~\ref{tab:dataset}, we present a comparison of existing datasets for PC-based mmWave human sensing, originally designed for human pose estimation (HPE) task with human joints/keypoints annotations. We focus on several key attributes, including mmWave radar sensor type, keypoint annotations, radar point cloud format, dataset size, and the variety of scenes covered in the dataset. Then, we further discuss the construction of HMP datasets from the selected datasets. 

\subsubsection{mmBody~\cite{chen2022mmbody}.} \textbf{Description:} The mmBody dataset is a public dataset for mesh reconstruction with multi-modal sensors: depth camera, RGB camera, and mmWave radar. Specifically, Phoenix mmWave radar from Arbe Robotics is chosen as the mmWave sensor, which extracts thousands of radar points for scene detection. Still, the detected radar point clouds are noisy and sparse compared to RGB and depth sensors. The dataset contains daily-life motions, with heterogeneous human motions including the motion of the torso, legs, arms, etc. Due to numerous subtle limb motions in the dataset, it is challenging to accurately predict human poses. To evaluate the robustness of mmWave HPE, the dataset includes various cross-domain scenes (lab2 and furnished) and adverse scenes (dark, rain, smoke, and occlusion).  Meanwhile, except for lab2 containing seen subjects, all other scenes contain unseen subjects for testing, which is challenging for the model's generalizability. Our experiment is conducted following the mmBody train-split settings~\cite{chen2022mmbody}. 

\begin{figure}[!t]
\centering
\includegraphics[width=3.0in]{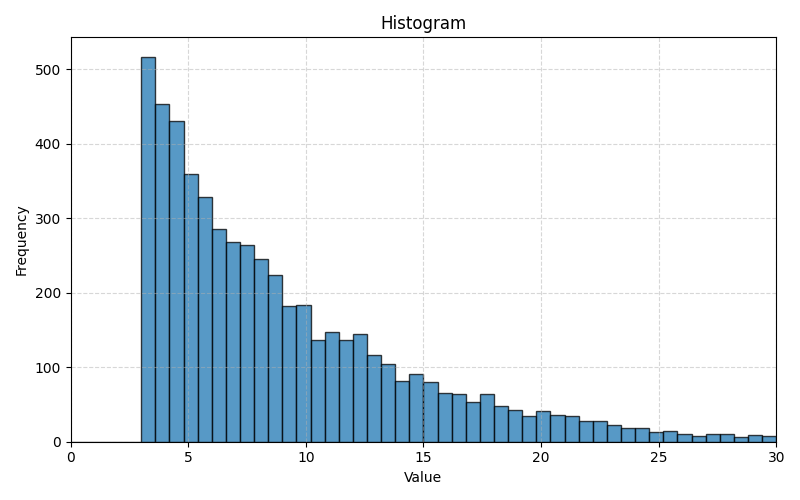}
\caption{Motion flow energy distribution after motion sample post-processing on the mmBody. Energy value $E_{flow} < thre = 3$ is considered as still motion and filtered.}
\label{mmBody_hist}
\end{figure}

\textbf{Motion sample selection:} A naïve approach to constructing an HMP dataset is to aggregate all $H+F$ frames of poses into a motion sample. However, the mmBody dataset contains many samples with minimal limb movement—e.g., only the head moves while the rest of the body remains largely static. Including such low-motion samples biases the model toward predicting static poses. To address this, we adopt an energy-based sample selection strategy in the DCT domain. Specifically, we first normalize each $H+F$-frame motion sample $x^{1:H+F}$ by subtracting the first frame $x^{[1]}$, yielding $x_{\text{norm}}^{1:H+F} = x^{1:H+F} - x^{[1]}$, which removes the influence of the initial pose and captures pure motion flow. The normalized sequence is then transformed into the DCT domain, and the motion energy is computed as $E_{\text{flow}} = \sum_{i=1}^{H+F} |x_{\text{norm}}^i|^2$. This energy reflects the overall motion magnitude, being larger for samples with more dynamic movement. Samples with $E_{\text{flow}} < \text{thre} = 3$ are discarded as they represent overly subtle motions (see Figure~\ref{mmBody_hist}). 

\begin{figure}[!ht]
\centering
\includegraphics[width=2.0in]{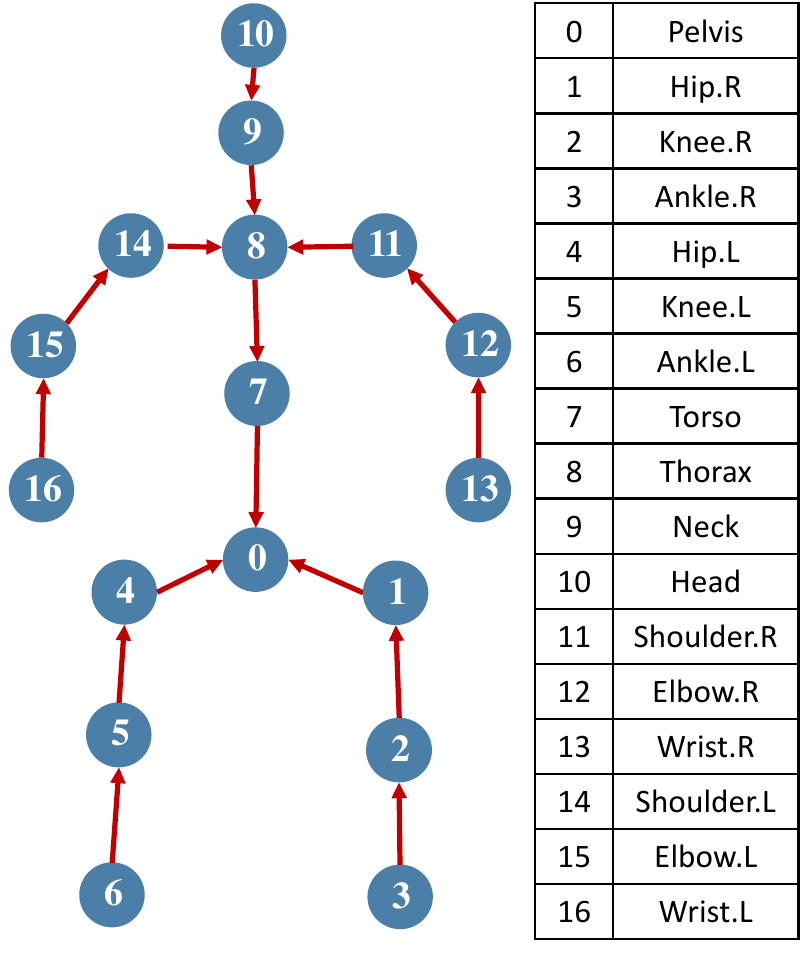}
\caption{Selected keypoints (ID and names) for mmWave human pose estimation. Red arrows indicate the select limbs.}
\label{pose}
\end{figure}

\subsubsection{mm-Fi~\cite{yang2023mm}.} mm-Fi offers a broader scope of human sensing, including action recognition and HPE, leveraging a variety of multi-modal sensors: RGB(D), LIDAR, WiFi, and mmWave radar. It is a large-scale dataset with 40 subjects participating and over 15k frames for training and testing. The mmWave radar utilized in the dataset is IWR6843 60-64GHz mmWave, which is a low-cost option generating a limited number of radar points. The point cloud format omits redundant range features. Meanwhile, different from the other two datasets, the model is trained in a self-supervised manner, as the human pose annotations are obtained by RGB images using HRNet-w48~\cite{prabhakara2023high}. The annotations are rather unstable compared to the motion capture systems. To test the model's generalizability, the dataset also proposes the cross-subject and cross-environment splits. Our experiment follows protocol 1 (P1) to include all daily-life activities and adopt only random split methods. 

\textbf{Motion sample selection:} Similarly to mmBody, we concatenate $H+F$ frames of poses as motion samples. No extra filter is required for mm-Fi.

\subsection{Preprocessing}

\subsubsection{Radar point clouds preprocessing.} Due to radar sparsity and the occasional miss-detection, we follow \cite{an2022fast} to apply a sliding-window strategy, concatenating adjacent frames to enrich the number of points. Specifically, 4 frames are concatenated for mmBody and 5 frames are concatenated for mm-Fi. Further, since mmWave radar point clouds are generated by the targets with salient Doppler velocity, the number of radar points is frame-wise variant. As a result, to enable mini-batch training using PyTorch dataloader~\cite{paszke2019pytorch}, we perform zero-padding (null points with 0 values) to guarantee the invariant input tensor shape. For mmBody, we zero-padding the point clouds to 5000 points, while a dynamic padding technique~\cite{yang2023mm} is applied for mm-Fi. Moreover, to handle noisy radar points resulting from environmental reflection and interference, we perform point cloud cropping to select only the points within the region of human activities. For mmBody, the region of human activities is centered at the ground-truth pelvis location, with a region size of ($x$:$\pm 1.6$m, $y$:$\pm 1.6$m, $z$:$\pm 1.6$m). However, for mm-Fi, as point clouds are generated solely by moving targets, cropping is unnecessary.

\subsubsection{Human pose preprocessing.} Our ground truth keypoints $H=\{h_1,..,h_{17}\}$ are selected according to Figure~\ref{pose}. Following ~\cite{ionescu2013human3} to construct ground-truth human poses, we perform pose normalization by pelvis alignment: subtracting the pelvis position $h_1$ from every keypoint $h_{1:17}$ of the skeleton.

\section{Supplementary of mmPred Implementations}

\subsection{Details of TPR} 
\begin{figure}[!t]
\centering
\includegraphics[width=3.0in]{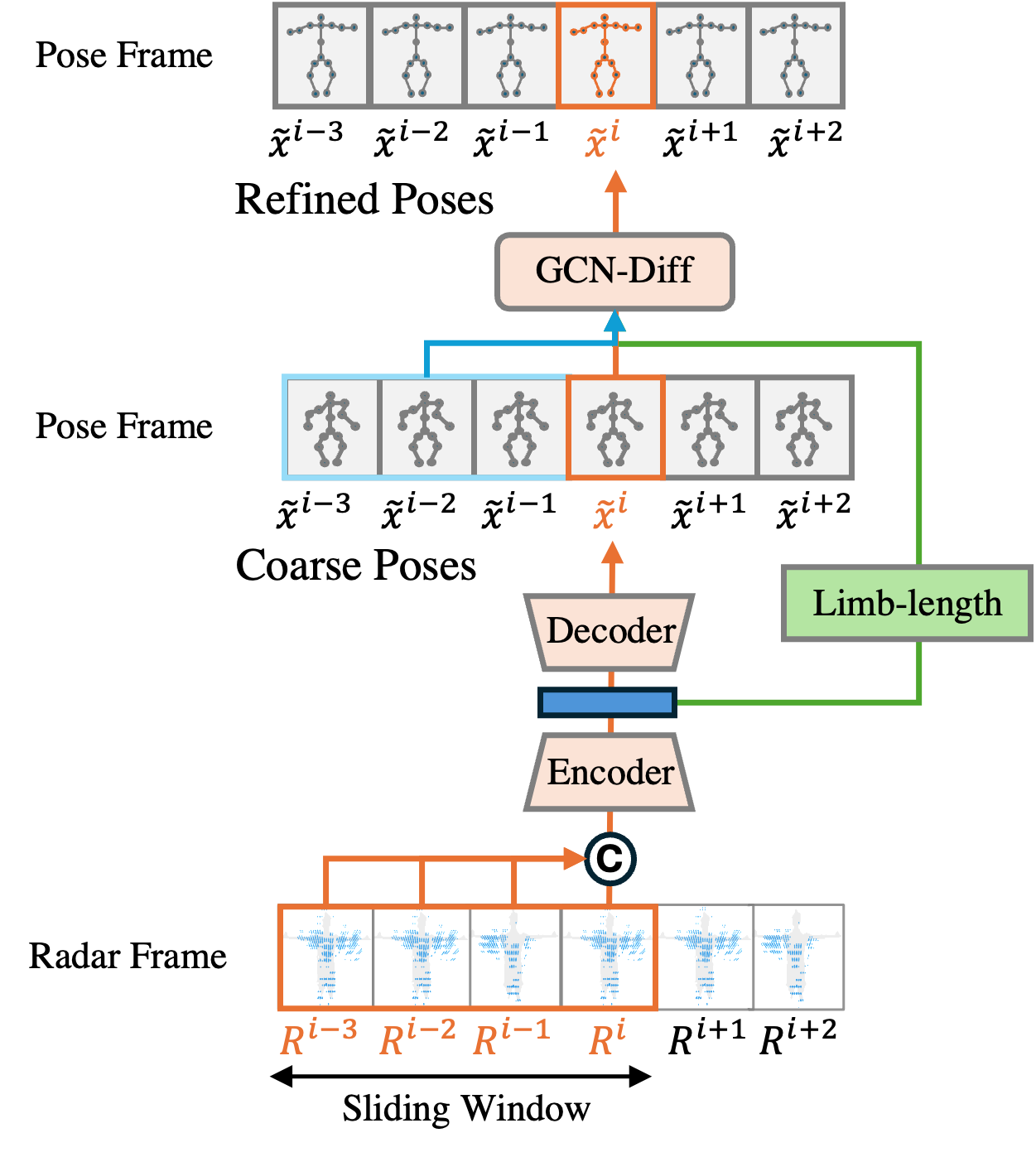}
\caption{Details of the Time-domain Pose Refining (TPR).}
\label{TPR}
\end{figure}

Given the sparsity of radar point clouds (PCs), we follow prior works~\cite{an2022fast, chen2022mmbody} and adopt a sliding-window strategy to enhance pose estimation. Specifically, for frame $i$, we concatenate radar PCs from the previous three frames $[R^{i-3}, R^{i-2}, R^{i-1}]$ with the current frame $R^i$ to form an aggregated input $\tilde{R}^i$. This multi-frame fusion is treated as the new radar PC representation for frame $i$, effectively serving as a temporally enriched single-frame observation (radar frame). The fused point cloud is then passed through a radar-based pose regressor—typically an encoder-decoder network. We employ Point4D~\cite{zhao2021point} and PointTransformer~\cite{fan2021point} as PC encoders, consistent with those used in our FDM module. These networks regress a sequence of coarse poses frame-by-frame.

To improve pose quality, particularly motion smoothness and anatomical plausibility, we introduce a diffusion-based pose refiner~\cite{fan2024diffusion}. The diffusion model takes the coarse pose as the initial state and progressively denoises it with five reverse steps. To guide the refinement, we condition the model on two information: (1) a short sequence of previously predicted poses to ensure motion continuity, and (2) estimated limb lengths derived from radar features to enforce anatomical consistency. The refined poses exhibit enhanced temporal coherence (reduced jitter) and improved limb-length consistency\cite{fan2024diffusion}.

\subsection{Detail design of GST} 

\begin{figure}[!t]
\centering
\includegraphics[width=3.0in]{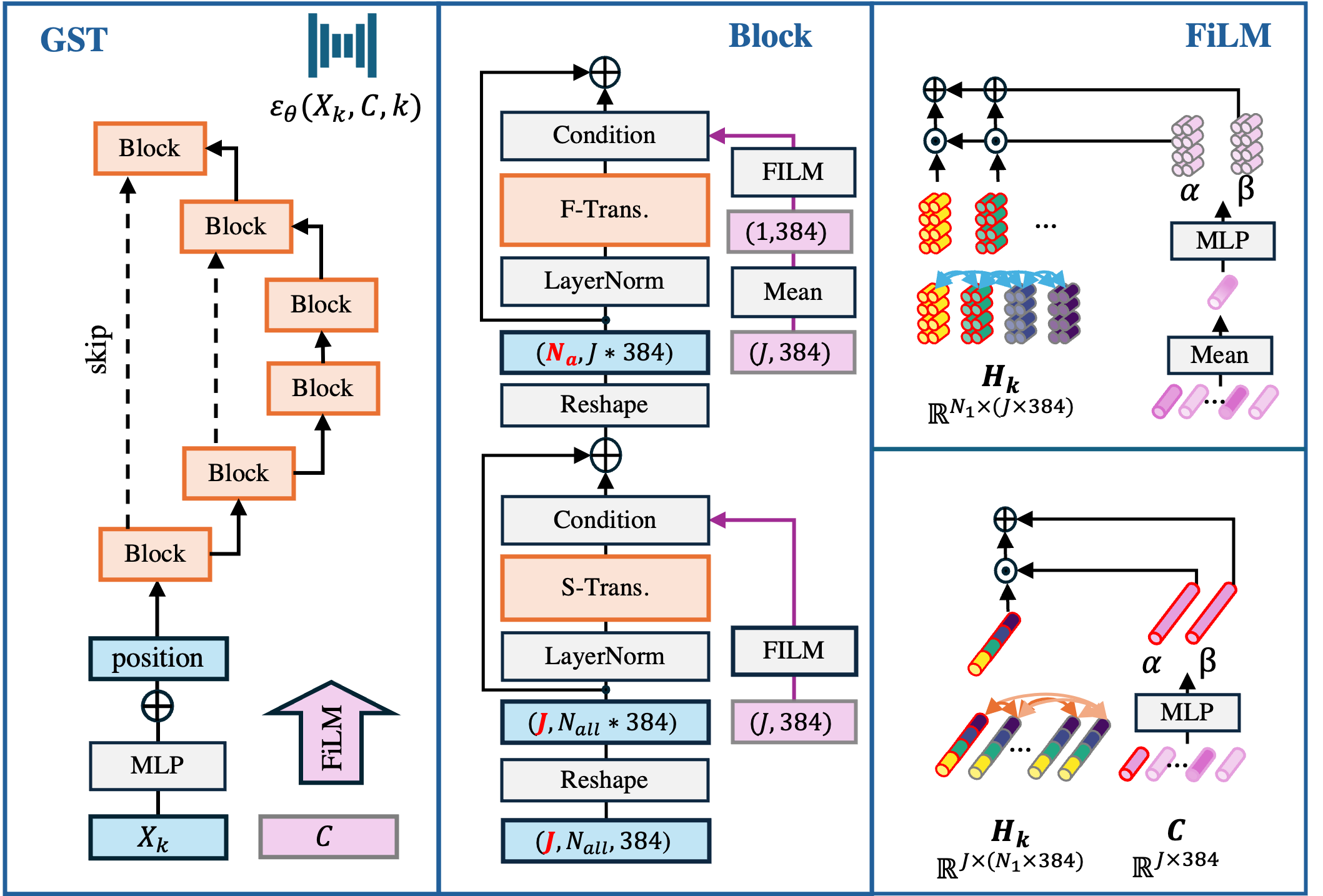}
\caption{Details of the GST architecture.}
\label{TPR}
\end{figure}

The GST module takes two inputs: the intermediate diffusion motion $X_k \in \mathbb{R}^{N_1 \times J \times 3}$ and the conditional embedding $C \in \mathbb{R}^{J \times 384}$. First, $X_k$ is mapped into the feature space via an MLP, yielding $H_k \in \mathbb{R}^{N_1 \times J \times 384}$, which contains $N_1 \times J$ tokens. To provide each token with awareness of its corresponding frequency and joint identity, we add a learnable, zero-initialized positional embedding of the same shape as $H_k$.

The resulting feature $H_k$ is then passed through a U-Net-style architecture, where skip connections are applied between encoder and decoder blocks. Each block consists of a Spatial Transformer (S-Transformer) and a Frequency Transformer (F-Transformer), which perform self-attention along the joint and frequency dimensions, respectively, as described in the main paper. Specifically, $H_k$ is reshaped to $\mathbb{R}^{J \times (N_1 \times 384)}$ for the S-Transformer and to $\mathbb{R}^{N_1 \times (J \times 384)}$ for the F-Transformer.

Within each transformer, we apply Feature‑wise Linear Modulation (FiLM)-based conditioning. For the S-Transformer, the conditional embedding $C \in \mathbb{R}^{J \times 384}$ is projected via an MLP to generate scaling and shifting parameters $\alpha, \beta \in \mathbb{R}^{J \times (N_1 \times 384)}$. Conditioning is then applied independently to each joint token $H_k^j \in \mathbb{R}^{1 \times (N_1 \times 384)}$ using the FiLM operation:

\begin{equation}
H_k^{j'} = H_k^j \odot \alpha^j + \beta^j, \quad j \in [1, \ldots, J].
\end{equation}
For the F-Transformer, we first apply mean pooling over the joint dimension of $C$, resulting in a global embedding $C' \in \mathbb{R}^{1 \times 384}$. This is further projected via an MLP to obtain $\alpha, \beta \in \mathbb{R}^{1 \times (J \times 384)}$, which are broadcasted across all $N_1$ frequency tokens. Conditioning is then applied to modulate the frequency-specific features.

\begin{figure}[!t]
\centering
\includegraphics[width=3.0in]{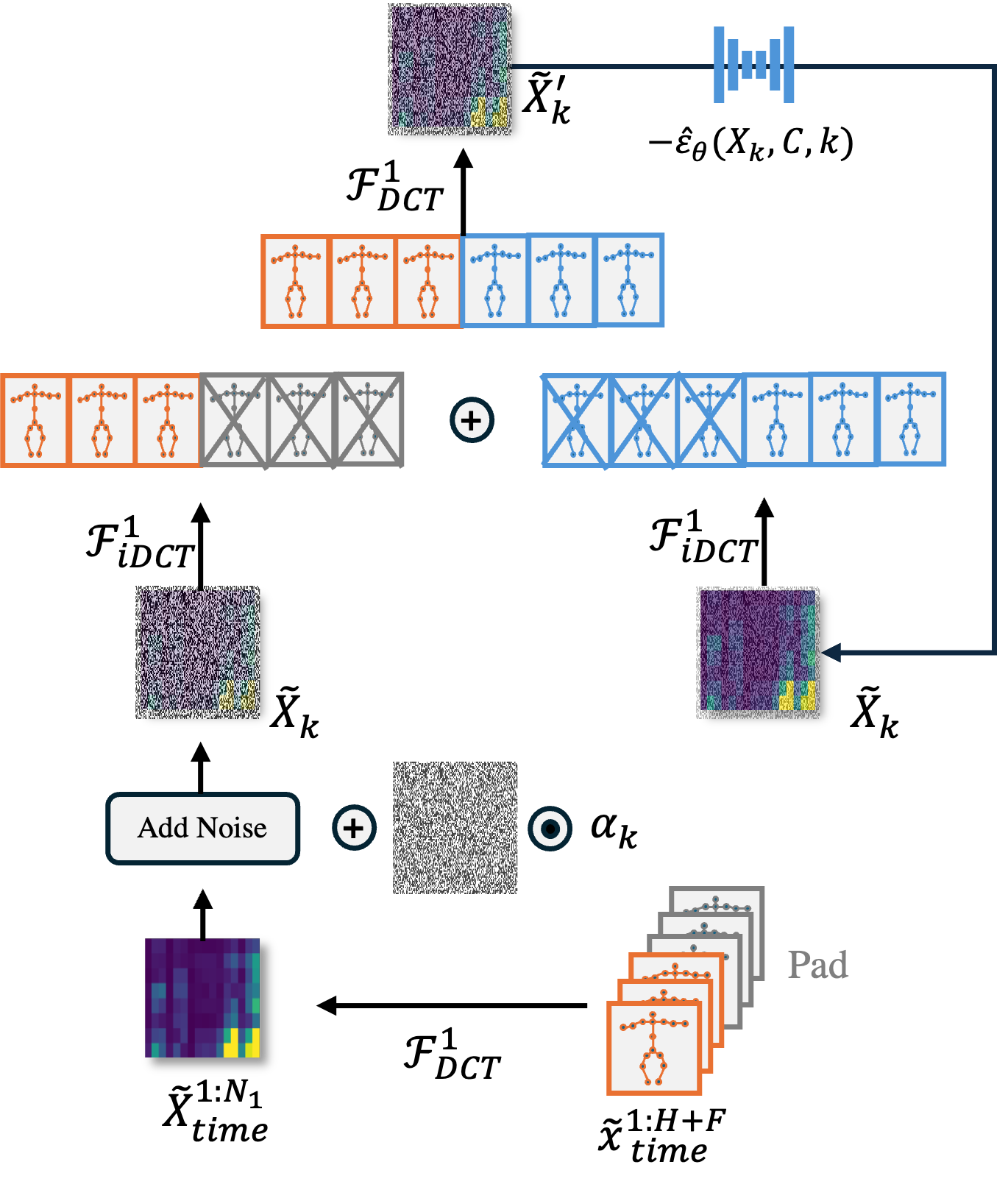}
\caption{Details of the sampling process of the reverse diffusion-based motion prediction.}
\label{reverse}
\end{figure}

\subsection{Detail design of diffusion reverse sampling process} 
This section discusses how we apply the pretrained diffusion model $\hat{\varepsilon}_\theta(X^k, k, C)$ for HMP via reverse sampling. To ensure that the generated motion respects the observed human motion history, we adopt the DCT-completion strategy proposed in HumanMAC~\cite{chen2023humanmac}, which treats human motion prediction as an inpainting task, preserving the historical segment while stochastically generating the future.

We first obtain the estimated historical motion $\tilde{x}^{1:H}_{\text{time}}$ (from radar PCs), perform padding, and transform it into the DCT domain. At each reverse diffusion step $k$, a noisy version of the historical motion is constructed as:
\begin{equation}
\tilde{X}_k = \tilde{X}^{1:N_1}_{\text{time}} + \mathcal{N}(0, \alpha_k^2),
\end{equation}
where $\alpha_k$ is the noise scale corresponding to diffusion step $k$. Simultaneously, the diffusion model outputs a predicted full motion $\hat{X}_k$ in the DCT domain. To perform inpainting, both $\tilde{X}_k$ and $\hat{X}_k$ are projected to the time domain via inverse DCT (iDCT), and then combined using a masking mechanism:
\begin{equation}
\hat{X}_k' = \mathcal{F}^1_{DCT} \left[ \mathbf{M} \odot \mathcal{F}^1_{iDCT}(\tilde{X}_k) + (1 - \mathbf{M}) \odot \mathcal{F}^1_{iDCT}(\hat{X}_k) \right],
\label{eq:dct-completion}
\end{equation}
where $\odot$ denotes the Hadamard product and $\mathbf{M} \in \mathbb{R}^{H+F}$ is the binary mask vector defined as:
\begin{equation}
\mathbf{M} = \left[
\underbrace{1, 1, \ldots, 1}_{H\text{-dim}},\;
\underbrace{0, 0, \ldots, 0}_{F\text{-dim}}
\right]^\top,
\end{equation}
ensuring that the history segment of $\hat{X}_k'$ strictly follows the estimated motion $\tilde{X}_k$. This completion mechanism stabilizes the reverse diffusion process and ensures that the generated motion is consistent with the observed past.

At the initial diffusion step $k = K$, we directly set the generated motion as the noised historical part, providing a meaningful initialization:
\begin{equation}
\hat{X}_K = \tilde{X}_{k}.
\end{equation}
Due to the cosine $\beta$-scheduling, the initial sample $\hat{X}_K$ is nearly pure noise.

\begin{figure*}[!t]
\centering
\includegraphics[width=.8\linewidth]{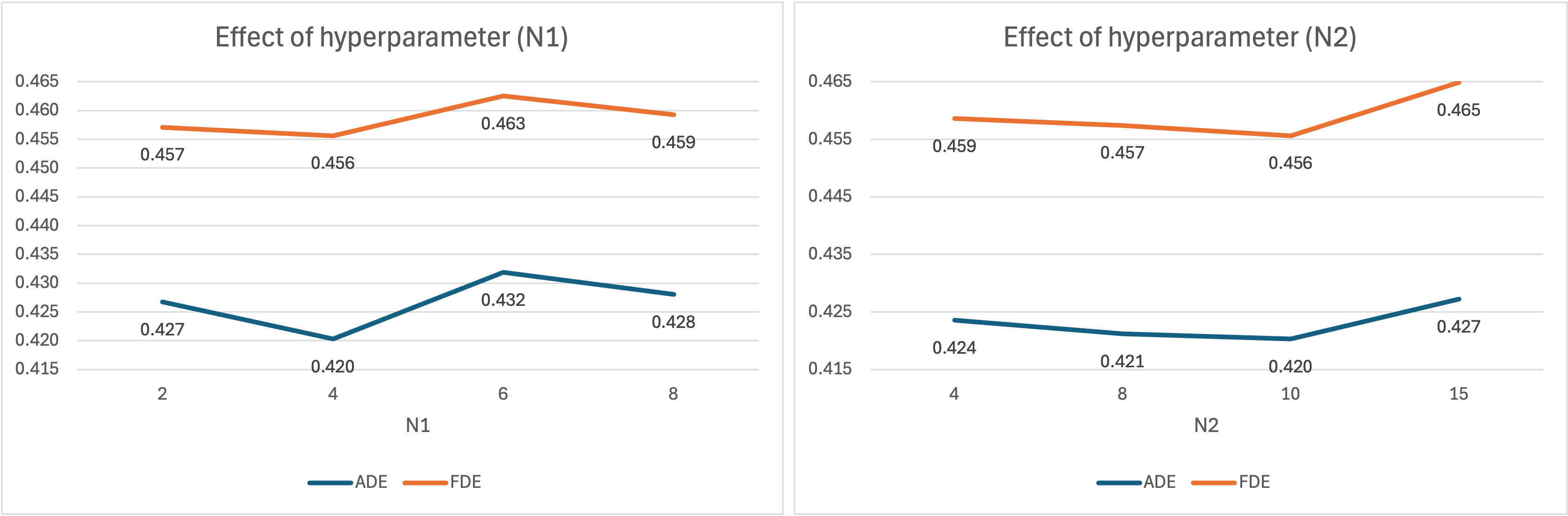}
\caption{Parameter sensitivity analysis of the number of DCT coefficients: $N_1$ for FDM and $N_2$ for the diffusion model.}
\label{hyperN} 
\end{figure*}

 \begin{table*}[!t]
\footnotesize
\centering
\caption{Evaluation with full stochastic HMP metrics. We report Average Pairwise Distance (APD) to quantify sample diversity, ADE/FDE to quantify prediction accuracy, and MultiModal ADE/FDE (MMADE/MMFDE) to measure prediction error averaged over multimodal futures.}
\resizebox{.6\textwidth}{!}{
  \fontsize{12}{14}\selectfont

    \begin{tabular}{cl|rrrrr}
    \toprule
    \multicolumn{2}{c|}{\multirow{3}[2]{*}{Method}} & \multicolumn{5}{c}{mmBody (Average)} \\
    \multicolumn{2}{c|}{} & \multicolumn{1}{c}{Diversity} & \multicolumn{2}{c}{Accuracy} & \multicolumn{2}{c}{MultiModal} \\
    \multicolumn{2}{c|}{} & \multicolumn{1}{l}{APD} & \multicolumn{1}{l}{ADE} & \multicolumn{1}{l}{FDE} & \multicolumn{1}{l}{MMADE} & \multicolumn{1}{l}{MMFDE} \\
    \midrule
    \midrule
    VAE   & DLow  & 1.315 & 0.484 & 0.519 & 0.656 & 0.586 \\
    \multirow{5}[0]{*}{Diffusion} & TCD   & 1.426 & 0.510 & 0.554 & 0.514 & 0.546 \\
          & Belfusion & 1.361 & 0.493 & 0.517 & 0.502 & 0.520 \\
          & Comusion & \textbf{1.697} & 0.486 & 0.515 & 0.492 & 0.515 \\
          & HumanMAC (radar) & 1.687 & 0.460 & 0.488 & 0.480 & 0.502 \\
          & HumanMAC (refine) & 1.625 & 0.456 & 0.487 & 0.474 & 0.499 \\
    Ours  & mmPred (all) & 1.540 & \textbf{0.421} & \textbf{0.457} & \textbf{0.451} & \textbf{0.481} \\
    \bottomrule
    \end{tabular}%
    }
    
    \label{tab:mmModality}

\end{table*}%

\section{Supplementary Results}
\label{sec:result}

\subsection{Full Stochastic HMP Evaluation}
In Tab.~\ref{tab:mmModality}, we report full stochastic HMP metrics following HumanMAC~\cite{chen2023humanmac}, together with additional stochastic baselines for a comprehensive evaluation of our method. We include the traditional VAE-based DLow~\cite{yuan2020dlow}, as well as diffusion-based approaches CoMusion~\cite{sun2024comusion} with time-frequency domain fusion, BeLFusion~\cite{barquero2023belfusion} with multi-stage latent diffusion, and TCD~\cite{saadatnejad2023generic} for incomplete history observation. We use Average Pairwise Distance (APD) to quantify sample diversity, and MultiModal ADE/FDE (MMADE/MMFDE) to measure prediction error averaged over multimodal futures. The multimodal set is constructed by grouping futures whose starting pose is close to the endpoint of the observed history.

Our proposed \textit{mmPred} achieves substantially lower multimodal prediction error than existing methods for both the current ground-truth future and all pose-similar futures. Its diversity is comparable to, though slightly lower than, state-of-the-art diffusion-based approaches. This suggests that \textit{mmPred} does not rely on random guessing over possible futures, but instead produces more accurate predictions that concentrate around the true future.

\subsection{Visualization of attention matrix in GST}

\begin{figure}[!t]
\centering
\includegraphics[width=.8\linewidth]{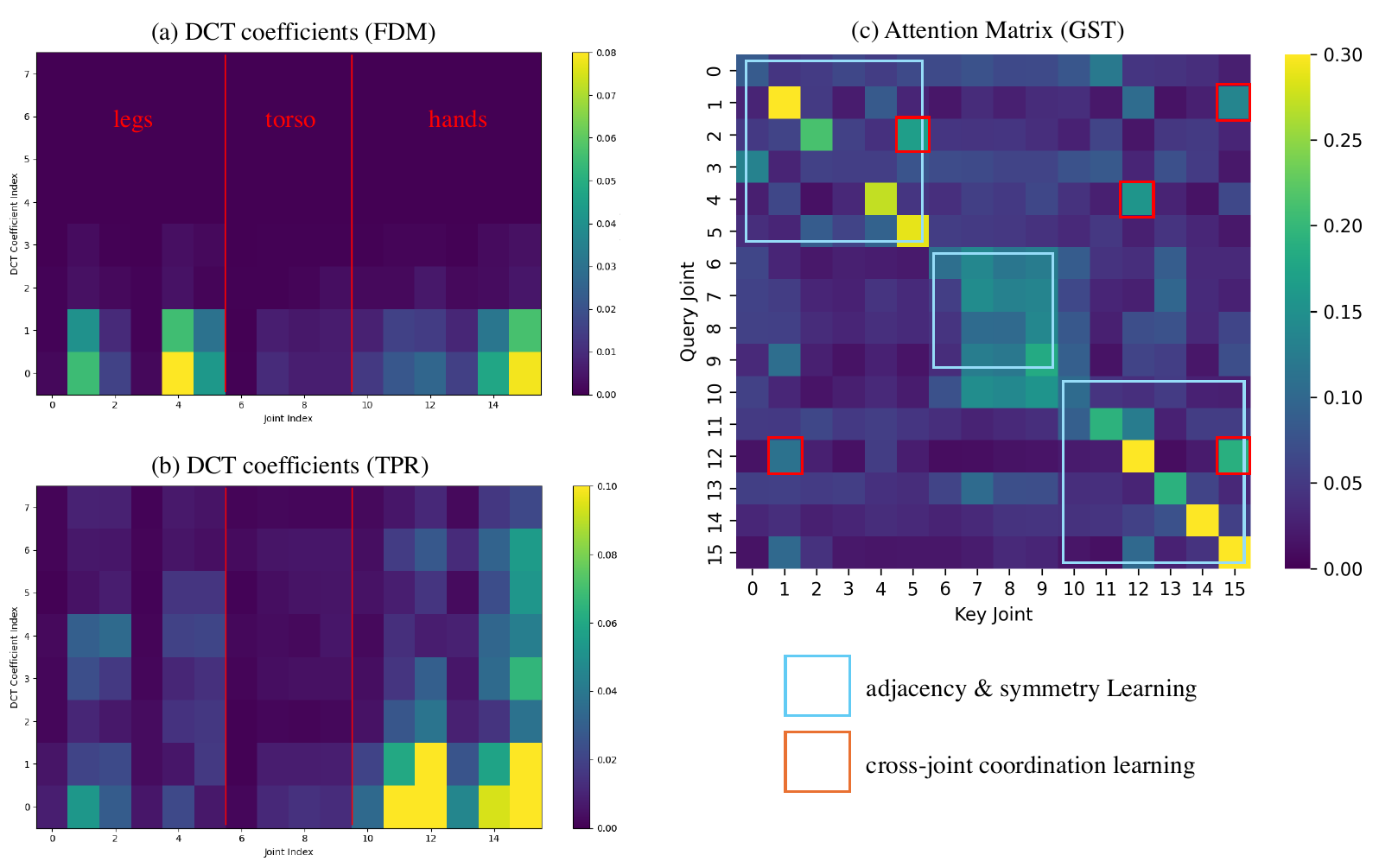}
\caption{(a) and (b) compare the predicted historical pose sequence from FDM and TPR in the DCT domain. Higher DCT values indicate more significant joint motion. Joint numbers (0-5), (6-9), and (10-15) belong to legs, torso, and hands, respectively. (c) plot the mean attention matrix during the reverse diffusion process.}
\label{GST-vis} 
\end{figure}

Figure~\ref{GST-vis}(a) and (b) first visualize the historical pose sequences in the DCT domain.
FDM concentrates on low-frequency components, indicating smoother motion trends, while the time-domain estimation exhibits more high-frequency energy, corresponding to joint vibration and noise.
Furthermore, in Figure~\ref{GST-vis} (c), we visualize how GST affects the joint feature extraction through joint-wise self-attention. Specifically, joint features self-attend more when FDM produces higher DCT responses (detected), while miss-detected joints (e.g., [0,3,10,13]) would aggregate features from others. This forms dynamic and global joint-wise dependencies, making feature learning more resilient to keypoint miss-detection (i.e., exploiting information from correlated joints). From the attention matrix, we observe both local adjacency and symmetry patterns (blue boxes), and long-range coordination such as hand–leg interaction (orange boxes).

 \begin{table*}[!t]
\footnotesize
\centering
\caption{Module efficiency evaluation on the mmBody dataset. The diffusion-based GST and the pose refiner are for one diffusion step.}
\resizebox{.8\textwidth}{!}{
  \fontsize{12}{14}\selectfont

    \begin{tabular}{l|cc|ccc}
    \toprule
    Modules & Input & Input Size & Latency (ms) & \#Param. & GFLOPs \\
    \midrule
    \midrule
    Pose Regressor (TPR) & $R^i$ & $\mathbb{R}^{1\times N \times 6}$ & 40.48 & 128.00   & 43.5 \\
    Pose Refiner (TPR) & $f_{pose}(R^{1:H})$ & $\mathbb{R}^{H \times J \times 3}$ & 11.85 & 18.51 & 0.62 \\
    FDM   & $R^{1:H}$ & $\mathbb{R}^{H\times N \times 6}$ & 41.44 & 128.71 & 81.02 \\
    GST (w/ Fusion) & $\tilde{x}^{1:H}_{time}, \tilde{X}^{1:N_2}_{freq}$ & $\mathbb{R}^{H\times J\times 3}, \mathbb{R}^{N_2\times J\times 3}$ & 8.13 & 9.59  & 0.10 \\
    \bottomrule
    \end{tabular}%
    }
    
    \label{tab:efficiency}

\end{table*}%

\begin{table}[ht]
\footnotesize
\centering
\caption{Parameter sensitivity analysis of the diffusion DDIM steps. We record the prediction accuracy by ADE and FDE, and motion realism by limb error and limb jitter. Bold is the best.}

\resizebox{.7\linewidth}{!}{

    \begin{tabular}{c|cc|cc}
    \toprule
    \multirow{2}[2]{*}{DDIM Steps} & \multicolumn{2}{c|}{Accuracy} & \multicolumn{2}{c}{Realism} \\
          & ADE   & FDE   & Error & Jitter \\
    \midrule
    \midrule
    20    & 0.419 & 0.454 & 7.52  & 1.19 \\
    40    & 0.420 & 0.453 & 7.38  & 1.21 \\
    60    & 0.418 & \textbf{0.452} & \textbf{7.29} & \textbf{1.20} \\
    80    & \textbf{0.417} & 0.453 & 7.30  & 1.24 \\
    100   & 0.420 & 0.456 & 7.36  & 1.26 \\
    \bottomrule
    \end{tabular}%

    }
    
    \label{tab:sense_K}

\end{table}%

\begin{table}[ht]
\footnotesize
\centering
\caption{Parameter sensitivity analysis of the number of GST layers. We record the prediction accuracy by ADE and FDE, and motion realism by limb error and limb jitter. Bold is the best.}

\resizebox{.7\linewidth}{!}{

    \begin{tabular}{c|cc|cc}
    \toprule
    \multirow{2}[2]{*}{GST layer} & \multicolumn{2}{c|}{Accuracy} & \multicolumn{2}{c}{Realism} \\
          & ADE   & FDE   & Error & Jitter \\
    \midrule
    \midrule
    2     & 0.427 & 0.465 & 7.39  & 1.18 \\
    4     & \textbf{0.420} & 0.458 & 7.36  & 1.26 \\
    6     & 0.421 & \textbf{0.452} & \textbf{7.36} & \textbf{1.23} \\
    8     & 0.422 & 0.456 & 7.43  & 1.30 \\
    \bottomrule
    \end{tabular}%

    }
    
    \label{tab:sense_layer}
\end{table}

\subsection{Efficiency analysis of mmPred} 
We present the efficiency analysis in Table~\ref{tab:efficiency}, demonstrating that our proposed modules require minimal computational resources during both historical motion estimation (TPR, FDM) and diffusion-based future prediction (GST). Notably, the existing pose regressor P4Transformer~\cite{fan2021point}, used as a baseline in our work, performs frame-by-frame pose estimation. In contrast, our FDM module directly processes all-frame historical radar point clouds and outputs a compact frequency-domain motion representation. Furthermore, FDM operates in parallel with TPR, enabling efficient dual-domain-representation extraction. Additionally, the proposed GST operates in the frequency domain and requires only 8ms per diffusion iteration—substantially faster than prior diffusion-based methods that rely on time-domain processing with additional autoencoding overhead. Moreover, our modules have a lightweight design in terms of parameter count and GFLOPs, especially when compared to the radar-based pose regressor~\cite{fan2021point}, further demonstrating suitability for resource-constrained scenarios such as robotics, edge computing, and IoT applications.

\subsection{Effect of hyper-parameters}
Overall, our proposed model is not highly sensitive to hyperparameter settings. Nevertheless, we discuss the influence of several key hyperparameters and their impact on performance.

\subsubsection{Effect of DDIM steps.} As presented in Table~\ref{tab:sense_K}, we observe that both overly large and small numbers of DDIM sampling steps negatively impact performance. A large number of DDIM steps may lead to increased stochasticity in the generated motion, while too few steps result in insufficient refinement during the reverse process. Empirically, we find that using 60 DDIM steps yields the best trade-off performance. Meanwhile, reducing DDIM steps also improves the model efficiency.

\subsubsection{Effect of the DCT coefficients number $N_1$ and $N_2$.} As shown in Figure~\ref{hyperN}, using either too few or too many DCT coefficients can degrade performance. A small number of coefficients may discard essential motion information, while an excessively large number may cause the model to overfit high-frequency components, leading to jittery motion predictions.

\subsubsection{Effect of number of GST layers.} As shown in Table~\ref{tab:sense_layer}, the number of layers should not be too large or too small. An excessive number of layers can lead to overfitting and reduced efficiency, while too few layers may result in underfitting and insufficient model capacity.


\end{document}